\documentclass{article}
\usepackage{authblk}
\usepackage[utf8]{inputenc}
\usepackage{bm}
\usepackage{tikz}
\usepackage{subcaption}
\usepackage{amsmath}
\usepackage{hyperref}
\usepackage{enumitem}

\usepackage[
natbib = true,
style = authoryear,
citestyle = authoryear,
maxcitenames = 1,
maxbibnames = 99,
giveninits=true, 
uniquename=false, 
uniquelist=false
]{biblatex}

\usepackage[margin=1in]{geometry}

\let\cite\textcite

\title{Deep Learning for Survival Analysis: A Review}
\author[1,3,4]{Simon Wiegrebe}
\author[2,3]{Philipp Kopper}
\author[5]{Raphael Sonabend}
\author[2,3]{Bernd Bischl}
\author[2,3]{Andreas Bender}
\affil[1]{Statistical Consulting Unit StaBLab, Department of Statistics, LMU Munich, Munich, Germany.}
\affil[2]{Department of Statistics, LMU Munich, Munich, Germany.}
\affil[3]{Munich Center for Machine Learning, LMU Munich, Munich, Germany.}
\affil[4]{Department of Genetic Epidemiology, University of Regensburg, Regensburg, Germany.}
\affil[5]{MRC Centre for Global Infectious Disease Analysis, Jameel Institute, Imperial College London, School of Public Health,
London, UK.}
\date{}                     %% if you don't need date to appear
\begin{document}

\maketitle

\begin{abstract}
\noindent The influx of deep learning (DL) techniques into the field of survival analysis in recent years has led to substantial methodological progress; for instance, learning from unstructured or high-dimensional data such as images, text or omics data. In this work, we conduct a comprehensive systematic review of DL-based methods for time-to-event analysis, characterizing them according to both survival- and DL-related attributes.
In summary, the reviewed methods often address only a small subset of tasks relevant to time-to-event data --- e.g., single-risk right-censored data --- and neglect to incorporate more complex settings.
Our findings are summarized in an editable, open-source, interactive table: \href{https://survival-org.github.io/DL4Survival}{https://survival-org.github.io/DL4Survival}. As this research area is advancing rapidly, we encourage community contribution in order to keep this database up to date.
\end{abstract}

\section{Introduction}
\label{Introduction}

\emph{Survival analysis} (SA), or equivalently \emph{time-to-event analysis}, comprises a set of techniques enabling the unbiased estimation of the distribution of outcome variables that are partially censored, truncated, or both. Usually, the outcome is given by the time until the occurrence of an event such as death, system failure, or time to remission.

Non-parametric methods like the Kaplan-Meier estimator \citep{kaplan1958nonparametric} are baseline tools still used today, yet semi-parametric methods received the most attention historically, in particular the Cox proportional hazards regression model \citep{cox1972regression} and its extensions.
% Both non-parametric and semi-parametric methods make no assumptions about the underlying distribution of event times and are popular baseline models in benchmark experiments as they are easy to fit and perform reasonably well \citep[see, e.g.,][for a survival benchmark on multi-omics data
% ]{herrmann2021large}. 
Since the early 2000s, Machine Learning (ML) methods have been successfully adapted to survival tasks: e.g., Random Survival Forest \citep{ishwaran2008random} and boosting-based methods \citep{binder2008allowing}. These methods often outperform traditional statistical models in terms of predictive power \citep{steele2018machine} (see \cite{wang2019machine} and \cite{Sonabend2021} for detailed discussions).

Neural networks (NNs) had already been applied to survival tasks in the 1990s \citep{faraggi1995neural, brown1997use}, but were shallow and restricted to the most standard survival settings. Most modern Deep Learning (DL) survival models have been developed only since the late 2010s, as indicated by the publication year of the methods we review; see \textit{Main Table} (\href{https://survival-org.github.io/DL4Survival}{https://survival-org.github.io/DL4Survival}). 
% They distinguish themselves from those first methods by being deep, employing more complex architectures, and covering a variety of different data settings.

Despite the large number of DL-based survival methods proposed in recent years, to the best of our knowledge, there is no general systematic review of these methods. \cite{schwarzer2000misuses} summarize misuses in early applications of NNs to clinical data. \cite{lee2019review} and \cite{deepa2022systematic} do not explicitly focus on DL-based survival methods, do not address any survival-related specifics of DL, and are restricted to use cases involving genomics data and cancer survival prediction, respectively. The glioma-focused survey by \cite{wijethilake2021glioma} as well as the benchmarking study by \cite{zhang2022survbenchmark} consider only few NN-based methods and thus do not provide a general overview of DL methods for time-to-event data either.
% A systematic review should help channel methodological efforts toward the most promising fields, direct research to areas with gaps, and also equip practitioners with a comprehensive overview of suitable methods for a given survival task.

Motivated by the above, in this paper we provide a comprehensive review of currently available DL-based survival methods, addressing theoretical dimensions, such as model class and NN architecture, as well as data-related aspects, such as outcome types and feature-related aspects (see Section \ref{DL in Survival Analysis} for definitions). Table \ref{tab:teaser} gives an overview of the dimensions we consider. 

% \begin{table*}[b]
% \tabcolsep=0pt%%
% \begin{tabular*}{\textwidth}{@{\extracolsep{\fill}}lll@{\extracolsep{\fill}}}
% % \toprule
% \hline
% \textbf{Dimension} & \textbf{Examples} & \textbf{Section(s)} \\
% % \midrule
% \hline
% estimation & Cox-based, discrete-time & section \ref{Estimation} \\ 
% neural network architecture & FFNN, CNN, RNN & sections \ref{Architectural Choices} and \ref{Network Architecture} \\ 
% outcome types & right-censoring, left-truncation, competing risks & sections \ref{Outcome Types} and \ref{Supported Outcome Types}\\ 
% feature-related aspects & time-varying features, high-dimensional features & sections \ref{Feature-related Aspects} and \ref{Supported Feature-related Aspects} \\ 
% % interpretability & classical statistical interpretability of tabular data, post-hoc methods & section \ref{Interpretability} \\ 
% % reproducibility & code and data freely accessible & section \ref{Reproducibility and Means of Evaluation} \\ 
% % means of evaluation & synthetic data, real-world data & section \ref{Reproducibility and Means of Evaluation} \\
% % \botrule
% \hline
% \end{tabular*}
% \caption{Overview of theoretical and practical dimensions reviewed.}
% \label{tab:teaser}
% \end{table*}

\begin{table*}[h]
\tabcolsep=0pt%%
\begin{tabular*}{\textwidth}{@{\extracolsep{\fill}}p{0.3\textwidth}p{0.4\textwidth}l@{\extracolsep{\fill}}}
\hline
\textbf{Dimension} & \textbf{Examples} & \textbf{Section(s)} \\
\hline
Estimation & Cox-based, discrete-time,\newline PEM-, ODE-, or continuous-time ranking-based & Section \ref{Estimation} \\ 
\hline
Neural network\newline architecture & FFNN, CNN, RNN, AE,\newline transformer, flexible, nODE & Sections \ref{Architectural Choices} and \ref{Network Architecture} \\ 
\hline
Outcome types & interval-, right- and left-censoring,\newline right- and left-truncation,\newline CR, MSM, recurrent events & Sections \ref{Outcome Types} and \ref{Supported Outcome Types}\\ 
\hline
Feature-related aspects & TVF, TVE, multimodality,\newline high-dimensional features & Sections \ref{Feature-related Aspects} and \ref{Supported Feature-related Aspects} \\ 
\hline
Interpretability & inherent interpretability,\newline post-hoc methods & Section \ref{Interpretability} \\ 
\hline
\end{tabular*}
\caption{Overview of theoretical and practical dimensions reviewed.\newline PEM: Piecewise Exponential Model; ODE: Ordinary Differential Equation; FFNN: Feed-forward Neural Network; CNN: Convolutional Neural Network; RNN: Recurrent Neural Network; AE: Autoencoder; nODE: neural ODE; CR: Competing Risks; MSM: Multi-state ; TVF: Time-varying Features; TVE: Time-varying Effects.}
\label{tab:teaser}
\end{table*}

This paper is structured as follows.
Section \ref{Theoretical Concepts and Data-related Aspects} introduces SA notation and concepts (Section \ref{Notation}), common data-related aspects of survival tasks (Section \ref{Data-related Aspects}), as well as estimation of survival models (Section \ref{sec:Estimation}).
Section \ref{DL in Survival Analysis} outlines the review methodology (Section \ref{inclusion_exclusion}), explains general NN architecture choices in SA (Section \ref{Architectural Choices}), and eventually provides a detailed, comprehensive overview of all methods reviewed, covering estimation and network architecture (Section \ref{Estimation and Network Architecture}), and supported survival tasks in terms of outcome types and feature-related aspects (Section \ref{Supported Survival Tasks})
% , 
% interpretability (Section \ref{Interpretability}) as well as reproducibility and means of evaluation (Section \ref{Reproducibility and Means of Evaluation})
; findings are summarized in the \textit{Main Table}. 
Finally, Section \ref{Conclusion} concludes, discusses limitations, and provides an outlook.

\section{Theoretical Concepts and Data-related Aspects}
\label{Theoretical Concepts and Data-related Aspects}

In this section, we first introduce quantities that are targets of estimation in SA and characterize the distribution of a random variable $T>0$. Later, we describe censoring and truncation, which need to be accounted for in order to estimate these quantities (see Section \ref{Data-related Aspects}).

\subsection{Targets of Estimation}
\label{Notation}

Initially, assume that $T$ is continuous. Let $f_T(t)$ and $F_T(t) := P(T \leq t)$ be its density and cumulative distribution function, respectively. 
Then, the survival function of $T$ is defined as
\begin{equation}
    S_T(t) := P(T>t) = 1 - F_T(t),
% \label{eq:survival_function}
\nonumber
\end{equation}
i.e., the probability of surviving beyond $t$. The hazard rate,
\begin{equation}
    h_T(t) := \lim_{\Delta t \to 0} \frac{1}{\Delta t} P(t \leq T < t + \Delta t | T \geq t) = \frac{f_T(t)}{S_T(t)},
\label{eq:hazard_rate}
\end{equation}
is the instantaneous risk of the event occurring given it has not yet occurred at time $t$. 
% Both survival function and hazard rate are often targets of estimation in SA and fully characterize the distribution of $T$. 
Finally, the cumulative hazard, defined as
\begin{equation}
    H_T(t) := \int_0^t h_T(u)du = -\log(S_T(t)),
% \label{eq:cumulative_hazard}
\nonumber
\end{equation}
is often used as an intermediate step when calculating the survival probability.

In the above, $T$ was assumed to be continuous. However, sometimes the time scale is discrete by nature (e.g., grade level at the time of school dropout) or a continuous time scale is discretized into intervals. 
% either because of substantive considerations (e.g., if only weekly or monthly information on event times is available or of interest) or for computational simplification
With discrete event times, the discrete hazard
\begin{equation}
    h_T(t) := P(T = t | T \geq t), \text{ } t = 1, 2, \dots
\label{eq:discrete_hazard}
\end{equation}
is the probability of the event occurring in the time interval $t$ conditional upon the individual still being alive at the beginning of $t$ (cf. \cite{tutz2016modeling} for details). This gives rise to the discrete-time survival probability
% \begin{equation}
    $S_T(t) := P(T>t) = \prod_{j=1}^t (1-h_T(j))$.
% \label{eq:discrete_survival_function}
% \nonumber
% \end{equation}
% i.e., the product of the complements of the discrete-time hazards across all time intervals (or time points) $j = 1, \dots, t$. 
% For further background on discrete time analysis from a statistical modeling point of view, see \cite{tutz2016modeling}.
Some discrete time methods on the other hand directly estimate the probability mass function (PMF), i.e. $P(T=t)$, rather than estimating the discrete hazard \eqref{eq:discrete_hazard}.

\subsection{Data-related Aspects}
\label{Data-related Aspects}

We now discuss different data-related aspects of time-to-event data, in terms of both outcomes and features, that are frequently encountered in real-world survival tasks. We refer to them as outcome types (Section \ref{Outcome Types}) and feature-related aspects (Section \ref{Feature-related Aspects}), respectively.

In Section \ref{Supported Survival Tasks}, we provide detailed information regarding which of the reviewed methods can handle these data-related aspects. 
% We will see that many methods do not immediately support settings beyond right-censored data. 

\subsubsection{Outcome Types}
\label{Outcome Types}

Throughout this work, we consider a sample of size $n$ and refer to a single $i \in \{1, \dots, n\}$ as \textit{individual} or \textit{subject}. Let $T_i > 0$ be the non-negative random variable representing the time until the event of interest for subject $i$ occurs.
We want to estimate the distribution of $T_i$ given the $p$-dimensional feature vector $\mathbf{x}_i$. 
However, $T_i$ often cannot be fully observed because the time-to-event is right-, left- or interval-censored. Let $C^L_i$ and $C^R_i$ be the left- and right-censoring times, and let $L_i$ and $R_i$ be the endpoints of the censoring interval for subject $i$, respectively. For an interval-censored observation, we have $T_i \in (L_i, R_i]$ as we only know that the event occurs within the interval, but not the exact time. Right-censoring $T_i \in (L_i = C^R_i, \infty]$ and left-censoring $T_i \in (L_i = 0, R_i = C^L_i]$ are special cases of interval-censoring.

Time-to-event data can also be subject to truncation. In SA, truncation implies that subjects are either not part of the dataset at all or not part of the risk set for a specific event at certain time points. 
% It can be a result of the selection criteria of a study or induced by the type of time-to-event process (primarily relevant in multi-state scenarios, including recurrent events). 
Formally, let $T^L_i$ and $T^R_i$ be the left- and right-truncation times, respectively. Left-truncation occurs when $T^R_i = \infty$, then subjects with $T_i < T^L_i$ never enter the study. Similarly, observations are right-truncated when $T^L_i=0$ and $T_i > T^R_i$. 
% This is often the case when data is sampled from registry data.

\begin{table*}[!ht]
\tabcolsep=0pt%%
% \begin{tabular*}{\textwidth}{@{\extracolsep{\fill}}lcc@{\extracolsep{\fill}}}
\begin{tabular*}{\linewidth}{@{\extracolsep{\fill}}p{0.25\linewidth}p{0.75\linewidth}@{}}
% \toprule
\hline
\textbf{Outcome type} & \textbf{Example} \\
% \midrule
\hline
% right-censoring & dropout in clinical trials: for some individuals, the exact event time is unobserved because they drop out of the study at some point \\ \hline
Right-censoring & clinical trials: exact event times are unobserved for some individuals because of dropout \\ \hline
Left-censoring & age at which children learn a certain task: some children already know the task at the beginning of the study, but it is unknown at which age they learned it \\ \hline
Interval-censoring & medical study with a periodic follow-up: exact event times are unknown, only the interval between two follow-ups is known \\ \hline
Right-truncation & transfusion-induced AIDS onset study \citep{klein_survival_1997}: only patients developing AIDS from transfusion before the registry sampling date are included, while patients with onset after that date are right-truncated \\ \hline
Left-truncation & coumarin abortion study \citep{meister2008statistical}: only women conscious of their pregnancy are included; women who had a spontaneous abortion before their pregnancy is recognized never enter the study \\ \hline
Competing risks & study on dialysis mortality \citep{noordzij2013we}: the event of interest, death on dialysis, is precluded by the competing event kidney transplantation \\ \hline
Multi-state & study on kidney failure: for all patients, transitions between the states \textit{healthy}, \textit{dialysis}, \textit{kidney transplantation} and \textit{death} are possible, sometimes even bidirectionally (e.g., between \textit{dialysis} and \textit{kidney transplantation}) \\ \hline
Recurrent event & incidence of pneumonia in young children \citep{ramjith2021flexible}: the occurrence of multiple, recurrent pneumonia episodes is possible, with episodes within a child's history not being independent \\
% \botrule
\hline
\end{tabular*}
\caption{Overview of different outcome types.}
\label{tab:outcome_modalities}
\end{table*}

Survival tasks are not restricted to single-risk scenarios. 
% Another outcome type is thus whether multiple, potentially competing events can occur. 
In case of \emph{competing risks}, each individual can experience only one of at least two distinct, mutually exclusive events, e.g., death in hospital versus hospital discharge.
More generally, in a \emph{multi-state} setting multiple (transient and terminal) events (states) are possible, as well as certain (recurring) transitions between them, e.g., transitions between different stages of an illness with death as terminal event. We denote transitions by $k\in \{1,\ldots,K\}$ and episodes by $e={1,\ldots,E}$.
% (if back-transitions are possible, some transitions could occur multiple times before a terminal event occurs).
A final outcome type we consider is \emph{recurrent events}. Often, we record a single outcome (censoring or event) for each individual. However, when conditions such as epilepsy or sports injuries are being modeled, subjects may experience the same event type repeatedly. 
% While recurrent events can be viewed as a special case of multi-state models, they can also be treated as standard survival data with subject-specific correlation \citep[e.g.,][]{box2006repeated}. 

Table \ref{tab:outcome_modalities} provides an overview and examples of the outcome types discussed in this section.

\subsubsection{Feature-related Aspects}
\label{Feature-related Aspects}

\emph{Time-varying features} (TVFs) such as weight or lifestyle factors change over time, whereas others such as sex are time-constant. Similarly, \emph{time-varying effects} (TVEs) are feature effects on the outcome (e.g., on the hazard rate) that vary over time. Both TVFs and TVEs constitute deviations from the proportional hazards (PH) assumption (see Section \ref{sec:Estimation}).

% Features such as weight or lifestyle factors may vary over time, whereas others such as sex are time-constant. 
% Similarly, feature effects on survival time may be time-dependent. \emph{Time-varying features} (TVFs) and \emph{time-varying effects} (TVEs) constitute deviations from the proportional hazards (PH) assumption (a very common assumption in SA, see Section \ref{sec:Estimation}) and, thus, need to be taken into account for survival modeling: with TVFs, the features themselves are observed at multiple time points and are hence time-dependent; with TVEs, the effect a feature has on the outcome (e.g., on the hazard rate or survival probability) varies over time. 

Another important feature-related aspect is the dimensionality of data input. Due to the prominence of SA in the life sciences, features derived from high-dimensional data --- \emph{omics} data in particular --- are sometimes employed to predict and explain survival times. In order for a method to learn from a high-dimensional feature space, the model architecture needs to be adapted, usually with appropriate penalization or feature selection techniques \citep[see, e.g.,][]{wu2019selective}.

\emph{Multimodality} is the final feature-related aspect we consider. In the life sciences, in particular, we are oftentimes not restricted to structured tabular data (e.g., clinical patient data), but also have access to unstructured data, such as images (e.g., CT scans) or text data (e.g., written doctor's notes); that is, the feature set is multimodal and special techniques are required to extract information from it.

\subsection{Estimation}
\label{sec:Estimation}
Here we summarize estimation in the SA context, focusing on the methods most frequently used among the DL-based approaches included in this review.
% thus omitting additive hazards models (XXX ref) and pseudo-value-based approaches (ref XXX). 

In SA we want to estimate the distribution of event times based on observed data, represented by tuples 

\begin{align}
(y^{entry}_{i,k,e}, y^{exit}_{i,k,e}, \delta_{i,k,e}, \mathbf{x}_{i,k,e}),
% \label{eq:tupel}
\nonumber
\end{align}

\noindent with $y^{entry}_{i,k,e}$ and $y_{i,k,e}^{exit}$ defining entry and exit times of subject $i=1,\ldots,n$ into the risk set for transition $k \in \{1,\ldots, K\}$ in episode $e = 1,\ldots E$, respectively, and $\delta_{i,k,e}\in \{0,1\}$ being the indicator for whether the respective transition has been actually observed (rather than censored). Finally $\mathbf{x}_{i,k,e}$ represents the $p$-dimensional feature vector (for simplicity we omit that $\mathbf{x}_{i,k,e}$ could additionally vary over time between the entry and exit times for transition $k$ in episode $e$). 
Often this notation can be simplified. For example, when all subjects enter the risk set at time point 0 and there is no truncation or interval-censoring, $y^{entry}_{i,k,e}$ is omitted. When we only consider one single event type, we can drop index $k$. If there are no recurrent transitions, we can additionally drop $e$, yielding the more common notation $(y_{i}, \delta_i, \mathbf{x}_i)$.

Parametric survival models, such as the Accelerated Failure Time (AFT) model \citep{kalbfleisch2011statistical}, assume event times to follow a certain statistical distribution characterized by a set of parameters. Based on the distribution-specific likelihood, parametric survival models then estimate these distributional parameters as a function of features $\mathbf{x}$. We can write the density for an event at time $t$ as

\begin{align}
   f(t|\boldsymbol{\theta}), & \text{ } t \geq 0,\\
    \boldsymbol{\theta} = \boldsymbol{\theta}(\mathbf{x})=(\theta_1(\mathbf{x}),\theta_2(\mathbf{x}),\ldots) &= (g_1(\mathbf{x}, \boldsymbol{\beta}_1), g_2(\mathbf{x}, \boldsymbol{\beta}_2), \ldots),
\label{eq:aft}
\end{align}

\noindent where $g_1(), g_2(), \dots$ are real-valued functions associating features $\mathbf{x}$ with the distributional parameters $\boldsymbol{\theta}(\mathbf{x})$ via parameters $\boldsymbol{\beta}_1,\boldsymbol{\beta}_2, \dots$. That is, all distributional parameters (e.g., both shape and scale of a Weibull distribution) can be estimated as a function of $\mathbf{x}$. Estimation proceeds by maximizing the likelihood given the observed data

\begin{align}
L(\boldsymbol{\theta}) = \prod_{i=1}^n L_i(\boldsymbol{\theta}) = \prod_{o \in \mathcal{O}} f(y_o)\times \prod_{c \in \mathcal{C}} S(y_c)\times \prod_{l \in \mathcal{L}}(1-S(y_l))\times \cdots,
\end{align}

\noindent where $L_i$ are the individual likelihood contributions, depending on the observed outcome type, and $\mathcal{O}$, $\mathcal{C}$, $\mathcal{L}$ are the sets of observed event times, right-censored, and left-censored observations, respectively. Likelihood contributions for other outcome types can be constructed similarly \citep[e.g.,][Ch. 3.5]{klein_survival_1997}.

Other methods exploit the relationships $f(t) = h(t)S(t)$ and $S(t) = \exp\left(-\int_0^t h(u)du\right)$, such that the likelihood can always be expressed in terms of only the hazard rate \eqref{eq:hazard_rate}, and right-censoring and left-truncation are dealt with by adjusting the so-called risk set 

\begin{equation}
\mathcal{R}(t) = \{(y_i^{entry}, y_i^{exit}, \delta_i, \mathbf{x}_i): y_i^{entry} < t \leq y_i^{exit}\}.
\label{eq:risk-set}
\nonumber
\end{equation}

\noindent Most prominently, the Cox PH regression \citep{cox1972regression} models the hazard rate at time $t$, conditional on features $\mathbf{x}$, as the product of a non-parametrically estimated baseline hazard $h_0(t)$ and the exponentiated log-risk $\eta = g(\mathbf{x}, \boldsymbol{\beta})$:

\begin{align}
h(t|\mathbf{x}) 
&= h_0(t)\exp(\eta = g(\mathbf{x}, \boldsymbol{\beta})).
\label{eq:cox_ph}
\end{align}

\noindent Feature effects are multiplicative with respect to the hazard rate independently of time, yielding proportionality of hazards.

Parameters are estimated by optimizing the log-partial-likelihood
\begin{align}
\hspace{-0.1cm}
Pl(\boldsymbol{\theta}) = 
\sum_{m=1}^M\,
\left(g(\mathbf{x}_{(m)}, \boldsymbol{\beta}) - \log\sum_{j \in \mathcal{R}(t_{(m)})} \exp\left(g(\mathbf{x}_{j}, \boldsymbol{\beta})\right) 
\right),
\label{eq:cox_loss}
% \nonumber
\end{align}

\noindent where $t_{(m)}$ is the $m$th ordered event ($m \in \{1, \dots, M\}$), $\mathcal{R}(t_{(m)})$ denotes the risk set at that time point, and $\mathbf{x}_{(m)}$ is the feature vector of the individual experiencing the event at $t_{(m)}$.

% Two further modeling approaches, described below, are \emph{reduction techniques} as they transform various survival tasks into regression or classification tasks by partitioning the follow-up into intervals and estimating the hazard or (conditional) survival probability in each interval. While some methods based on reduction techniques make distributional assumptions to estimate the parameters of the (interval-specific) hazard, they do not make assumptions about the distribution of event times, similar to the Cox model. Left-truncation and right-censoring are again dealt with by excluding subjects outside the risk set \eqref{eq:risk-set} for a specific interval.

Piecewise Exponential Models (PEMs) also parametrize the hazard rate as in \eqref{eq:cox_ph}. However, by partitioning the time axis into $J$ intervals and assuming piecewise constant hazards within each interval, the baseline hazard is parametrized and estimated alongside the feature-related coefficients. \cite{friedman1982piecewise} showed that the likelihood of this model is proportional to a Poisson likelihood, which implies that, after appropriate data transformation, any method capable of minimizing a negative Poisson log-likelihood can also be used for various survival tasks \citep{bender_general_2021}. Despite partitioning the follow-up into intervals, PEM-based approaches are methods for continuous time-to-event data as they take the full information about event times into account.

Discrete-time survival methods, such as discrete hazard methods \citep{tutz2016modeling} or Multi-Task Logistic Regression \citep[MTLR;][]{yu2011learning_2}, consider the time-to-event data to be a succession of binary outcomes. To do so, the time axis is first partitioned into intervals, with $T = t$ implying the event occurred in interval $(a_{t-1}, a_t]$. Binary event indicators $y_{it}$ are then defined for each time interval $t$ and used as outcomes. For individual $i$, the discrete hazard $h_i(t|\mathbf{x}_i)$ in interval $t$ is then
\begin{equation}
    h_i(t|\mathbf{x}_i) = \phi(g(\mathbf{x}, \boldsymbol{\beta})) = P(y_{it}=1|T \geq t, \mathbf{x}_i), 
    \label{eq:discrete}
\end{equation}

where the real-valued function $g()$ represents feature effects and $\phi()$ maps this quantity onto $[0,1]$ to yield the conditional event probability $P(y_{it}=1|T \geq t, \mathbf{x}_i)$.
Analogously to PEM-based approaches, any ML algorithm that is applicable to binary outcomes can be used for discrete-time survival modeling after data transformation. The logit model, for instance, uses a logistic response function to model the probability of the event taking place in $t$ (i.e., the discrete hazard \eqref{eq:discrete_hazard}), conditional on $a_{t-1} < t$ and feature values $\mathbf{x}$.
Alternatively, some discrete-time methods directly estimate the probability of an event at specified time points $P(y_{it} = 1|T=t,\mathbf{x}_i)$ using a softmax output layer. 
% A disadvantage of assuming time to be non-continuous is that interpolation between the limited, fixed prediction time points is necessary in order to achieve approximately continuous predictions. 
% Alternatively, one can choose a quasi-continuous grid of discrete event times which, however, may imply much higher computational complexity.\\

\section{Deep Learning in Survival Analysis}
\label{DL in Survival Analysis}

Early DL-based survival techniques date back to the mid-1990s \citep{liestbl1994survival, faraggi1995neural, brown1997use} and are usually NN-based extensions of classical statistical survival methods discussed in Section \ref{sec:Estimation}.
% \cite{faraggi1995neural} build upon the Cox PH model. 
While in the Cox model the log-risk \eqref{eq:cox_ph} is traditionally given by $g(\mathbf{x}, \boldsymbol{\beta}) = \mathbf{x}^\top \boldsymbol{\beta}$, the model by \cite{faraggi1995neural} replaces the linear predictor by a shallow feed-forward neural network (FFNN). \cite{liestbl1994survival} propose implementing the PEM as an NN, yet without any hidden layers. The \emph{PEANN} model by \cite{fornili2013piecewise} parametrizes the piecewise constant hazards by a shallow FFNN. \emph{PLANN} \citep{biganzoli1998feed} is an NN-based extension of the discrete-time logit model, parametrizing the discrete hazard by an FFNN.

Many DL-based methods for SA have been developed in recent years. They usually build upon one of the aforementioned statistical survival approaches, while harnessing advantages of NNs. Furthermore, recent advances in multimodal learning and interpretability have made DL-based survival methods even more attractive for many common survival tasks.

\subsection{Inclusion and Exclusion Criteria}
\label{inclusion_exclusion}

For this review, we designed a two-step literature screening process.
% , in line with the focus of this paper: providing an overview of distinct available methods and their capabilities, as opposed to summarizing (novel) applications of already existing techniques.
In the first step (inclusion criteria), we searched Web of Science for the topic 

\vspace{0.5cm}
\begin{minipage}{0.8\linewidth}
    \texttt{
        ("survival analysis" or "time-to-event analysis" or "survival data" or "time-to-event data") AND \\
        ("neural network" or "deep learning") AND \\
        ("model" or "method") AND \\
        ("performance" or "evaluation" or "comparison" or "benchmark")
    }
\end{minipage}
\vspace{0.5cm}

\noindent with December 31, 2022 as cutoff date. These inclusion criteria resulted in a total of 211 articles. In the second step (exclusion criteria), we excluded all articles not satisfying \emph{all} of the following four conditions:

\begin{itemize}
    \item[(a)] Development of a new DL-based method beyond the mere application of an already existing method to new data or contexts.
    \item[(b)] Evaluation of performance results on at least one non-private benchmark dataset.
    \item[(c)] Performance evaluation using metrics designed for time-to-event data, such as C-index or Integrated Brier Score.
    \item[(d)] Focus on estimation and prediction in the context of time-to-event data and learning all model parameters within the NN architecture in an end-to-end fashion.
\end{itemize}

\noindent Criteria (a), (b), and (c) aim to ensure that the paper in question develops a new method rather than applying a known method to new data or in a new context. Criterion (b) complements (a) as the predictive utility is often illustrated via benchmark experiments when new methods are proposed. Additionally, criterion (b) introduces an open science aspect and ensures that at least one empirical comparison could be replicated in theory. Criterion (c) ensures that benchmark analyses focus on methods modeling time-to-event data, as some papers that passed the initial screening eventually ignored the time-to-event nature of the data. Finally, criterion (d) aims to exclude two-step approaches where DL is used solely for feature extraction, with survival modeling performed using non-DL approaches with the extracted features outside of the NN. 
Our criteria are motivated by the scope of this work --- to review methods that can be used for specific time-to-event problems and to provide details on estimation-, architecture- and data-related aspects of the respective methods.
% We observe that more modern approaches are more likely to comply with these criteria, potentially as a result of a stronger standardization in the SA literature.
% However, earlier influential contributions like \cite{ranganath_deep_2016} or \cite{alaa2017deep} do not comply with our inclusion and exclusion criteria and are thus not within the scope of this analysis, despite belonging to the canonical literature.
% Up to this date, to our best knowledge, there has been no attempt to \emph{systematically} compare DL methods in SA, e.g., via a neutral benchmark study. 
% One such benchmark study does exist for the subdomain of SA with omics and high-dimensional data \citep{zhang2022survbenchmark}, showing that DL approaches do not perform very well; however, the authors only use default hyperparameter settings without tuning, which limits the generalizability of the study.
% Hence, we conclude that to this date conclusive benchmarking studies for DL in SA are lacking and should be addressed in future research.

\begin{figure*}[hb]
  % \centering
    \hspace{-3cm}
    \includegraphics[width=1.4\linewidth]{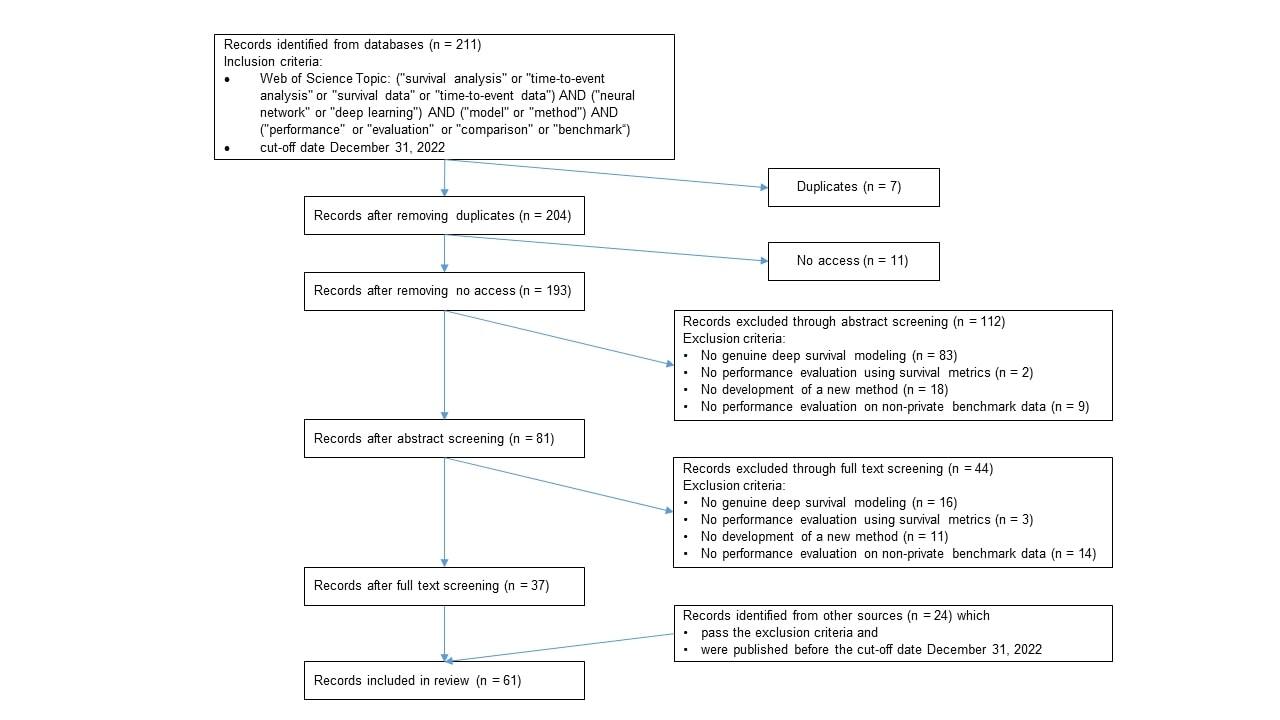}
  \caption{PRISMA diagram for literature screening of deep learning-based survival methods.}
  \vspace{-0.2cm}
  \label{fig:prisma}
\end{figure*}

Subsequently, we combined the selected articles with additional papers that had otherwise come to our attention and fulfilled the above criteria, yielding a total of 61 articles --- and thus, 61 distinct methods. The inclusion, exclusion, and screening process is visualized in the PRISMA diagram in Figure \ref{fig:prisma}.

The following naming scheme is used in the remainder of the paper to reference individual methods/papers: the method name as specified in the publication, if provided and unique; if the method name is not unique, we append a suffix (the first three letters of the first author's last name followed by the year of publication) with an underscore; if no method name is provided, we use this suffix as a name. All methods are summarized in our \textit{Main Table} (\href{https://survival-org.github.io/DL4Survival}{https://survival-org.github.io/DL4Survival}).

% Real-world survival tasks always come with their own idiosyncrasies, based on which the required capabilities of the survival methods are determined. For instance, a tabular medical dataset with multiple disease outcomes requires competing risk modeling but does not necessitate multimodal data techniques. 
We now aim to provide a summary of the 61 methods based on a broad range of both theoretical (estimation and architecture) as well as practical model characteristics (outcome types and feature-related aspects).

\subsection{Architectural Choices}
\label{Architectural Choices}

FFNNs were the earliest type of NN architecture \citep{ivakhnenko1967cybernetics, rosenblatt1967recent}. 
Within an FFNN, information passes from the input nodes through a user-specified number of hidden layers until the output nodes. 
Information only flows forward as there are no cyclical patterns or loops. The main property of FFNNs is stated through the universal approximation theorem (\cite{hornik1989multilayer}), meaning that NNs are capable of approximating a very general class of functions.
Practically, this allows FFNN to discover non-linear feature effects and complex interaction structures.
In SA, FFNNs naturally allow for a more flexible estimation of, e.g., (semi-)parametric hazard rates, as well as for the incorporation of TVEs and TVFs (in theory); for instance, the hazard rate in \eqref{eq:cox_ph} can be estimated more flexibly by parametrizing $g(\mathbf{x})$ through an NN. At the same time, the FFNN architecture contains multiple limitations: for example, learning from multimodal data input --- in particular, image data --- is not possible. FFNNs constitute the main architecture of most early DL-based survival methods and still serve as a baseline building-block within most advanced architectures.

Convolutional neural networks (CNNs) were introduced in the late 1980s \citep{lecun1989backpropagation} and are most successfully employed in computer vision. In time-to-event analysis, CNNs are usually applied to unstructured data, especially images. Often, CNN-based methods use large pre-trained CNNs with many parameters, such as ResNet18 \citep{he2016deep}, and then fine-tune them on case-specific data. This transfer learning approach enables the application of large CNNs to smaller datasets.

Recurrent neural networks (RNNs), also invented in the 1980s \citep{rumelhart1986learning}, distinguish themselves from FFNNs and CNNs by being able to memorize parts of the input through a short-term memory and are thus applicable to sequential data. In SA, RNNs are hence useful when TVFs are present or to take temporal information into account in general.

The autoencoder \citep[AE;][]{ballard1987modular} is another common NN architecture, learning how to reduce the dimensionality of input data and subsequently reconstructing the data from the learned latent representation; extensions include stacked AEs \citep[SAEs;][]{vincent2010stacked} and variational AEs \citep[VAEs;][]{Kingma2014}. 
General Adversarial Networks \citep[GANs;][]{goodfellow2014generative} consist of a generator that produces synthetic data of gradually improving quality as well as a discriminator that learns how to distinguish between true data input and generator-produced data points.
Transformers \citep{vaswani2017attention_2} use an attention mechanism to learn a representation of context in sequential (e.g., language) data and can subsequently produce output (sequences) from it.
Normalizing flows \citep[NFs;][]{rezende2015variational} constitute a family of generative models which employ differentiable and invertible mappings to obtain complex distributions from a simple initial probability distribution for which sampling and density evaluation is easy.
Neural Ordinary Differential Equations \citep[nODEs;][]{chen2018neural_2} use NNs to parametrize the derivative of the hidden state, thus moving beyond the standard specification of a discrete sequence of hidden layers. 
Fuzzy neural networks \citep{lee1975fuzzy} use fuzzy numbers as inputs and weights within the NN. 
Diffusion models \citep{sohl2015deep} employ a Markov chain to gradually add random noise to the input data and subsequently learn to undo this diffusion, learning to generate new data from noise.

Many adoptions of NNs for SA emphasize the replacement of the predictors in \eqref{eq:aft}, \eqref{eq:cox_ph}, or \eqref{eq:discrete} through a (deep) NN. The (DL-based) survival models can be further extended to also include interactions, non-linear effects, stratification, time-varying effects, and even unstructured components $d(\mathbf{z})$, yielding the generalized predictor 

\begin{align}
    \eta = g(\mathbf{x}, \mathbf{z}, t) = f(\mathbf{x}, t) + \gamma_1 d_1(\mathbf{z}_1) + \dots + \gamma_G d_G(\mathbf{z}_G),
\label{eq:predictor_extended}
\end{align}

where $f(\mathbf{x}, t)$ denotes potentially non-linear, time-varying effects of tabular features $\mathbf{x}$ as well as their interactions. $d_g(\mathbf{z}_g)$ denotes embeddings learned in the deep part(s) of the model from unstructured data sources $\mathbf{z}_g$, $g \in \{1,\dots,G\}$, such as images or text. That is, the predictor $g(\mathbf{x})$ from \eqref{eq:cox_loss} can be generalized to be $g(\mathbf{x}, \mathbf{z}, t)$. Using an appropriate transformation function $\psi$ predictor \eqref{eq:predictor_extended} can be transformed to e.g. the hazard function or cumulative incidence function, depending on the target of estimation.
% While this is associated with a more expressive and complex predictor, this approach does not tackle some of the existing problems with other feature-related aspects or outcome types.
% More recently, NNs have been shown to be particularly suited to facilitate specific modalities directly in their architecture.

Architectural choice is also closely related to parametrization. The PMF of discrete-time methods can be modeled via a softmax layer producing discrete survival probabilities at each (pre-defined) time point, as done in \cite{lee2018deephit}. 
RNN architectures are particularly suitable for taking into account temporal information and sharing parameters across time, e.g., in order to estimate quantities like the hazard rate or survival probability at time $t$ using information from time points $\tilde{t} < t$ \citep[e.g.,][]{giunchiglia_rnn-surv_2018}. 
Some less frequently encountered architectures, for example GANs, incentivize the development of custom losses \citep{chapfuwa2018adversarial}.
More recent work shows that (surrogate) loss functions can be created based on scoring rules, such as a smooth C-index loss function \citep{huang_deep_2018} or Survival-CRPS \citep{avati2020countdown}, for parameter estimation without requiring traditional inner loss functions like the negative log-likelihood.

\begin{figure*}[h]
  \centering
    \includegraphics[width=0.63\linewidth]{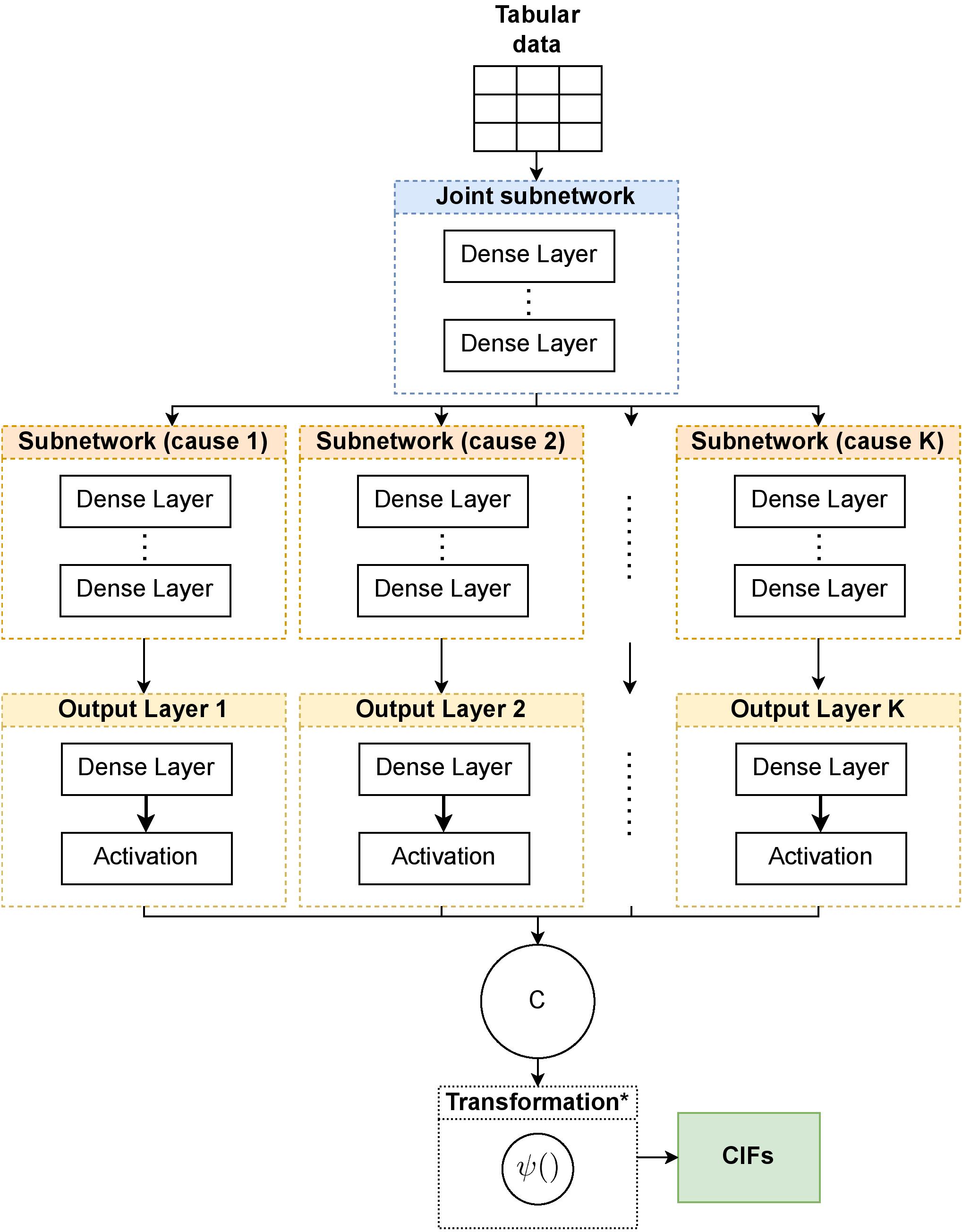}
  \caption{
  Schematic neural architecture for competing risks in survival analysis using shared and cause-specific subnetworks.
  % As a final step, the network provides a suitable activation function to match the loss.
  $\psi()$ transforms the model output (e.g., hazard rate) to the final outcome (e.g., cumulative incidence functions (CIFs)).
  % Depending on the kind of output that has been obtained directly from the model, we may need to transform the model output (e.g., from hazards) to the cumulative incidence functions (CIFs) with the function $\psi()$. In the case of a direct survival function estimation $\psi()$ is the identity function.
  }
  % \vspace{-0.5cm}
    \label{fig:architecture_CR}
\end{figure*}

It is furthermore possible to directly integrate some time-to-event data modalities into the architecture of deep survival models.
For example, shared and cause-specific subnetworks for cause- or transition-specific hazards in competing risks and multi-state modeling analysis via soft- or hard-sharing of parameters \citep{ruder2017overview} have been adopted by many DL-based survival methods when modeling transitions between different states (cf. Figure \ref{fig:architecture_CR}).
% Other methods represent time-to-event data as a longitudinal three-dimensional tensor format \citep[see, e.g.,][]{kopper2022deeppamm}, which results in the ability to capture time-dependence (second dimension) and competing risks (third dimension).

\begin{figure*}[h]
  \centering
    \includegraphics[width=0.63\linewidth]{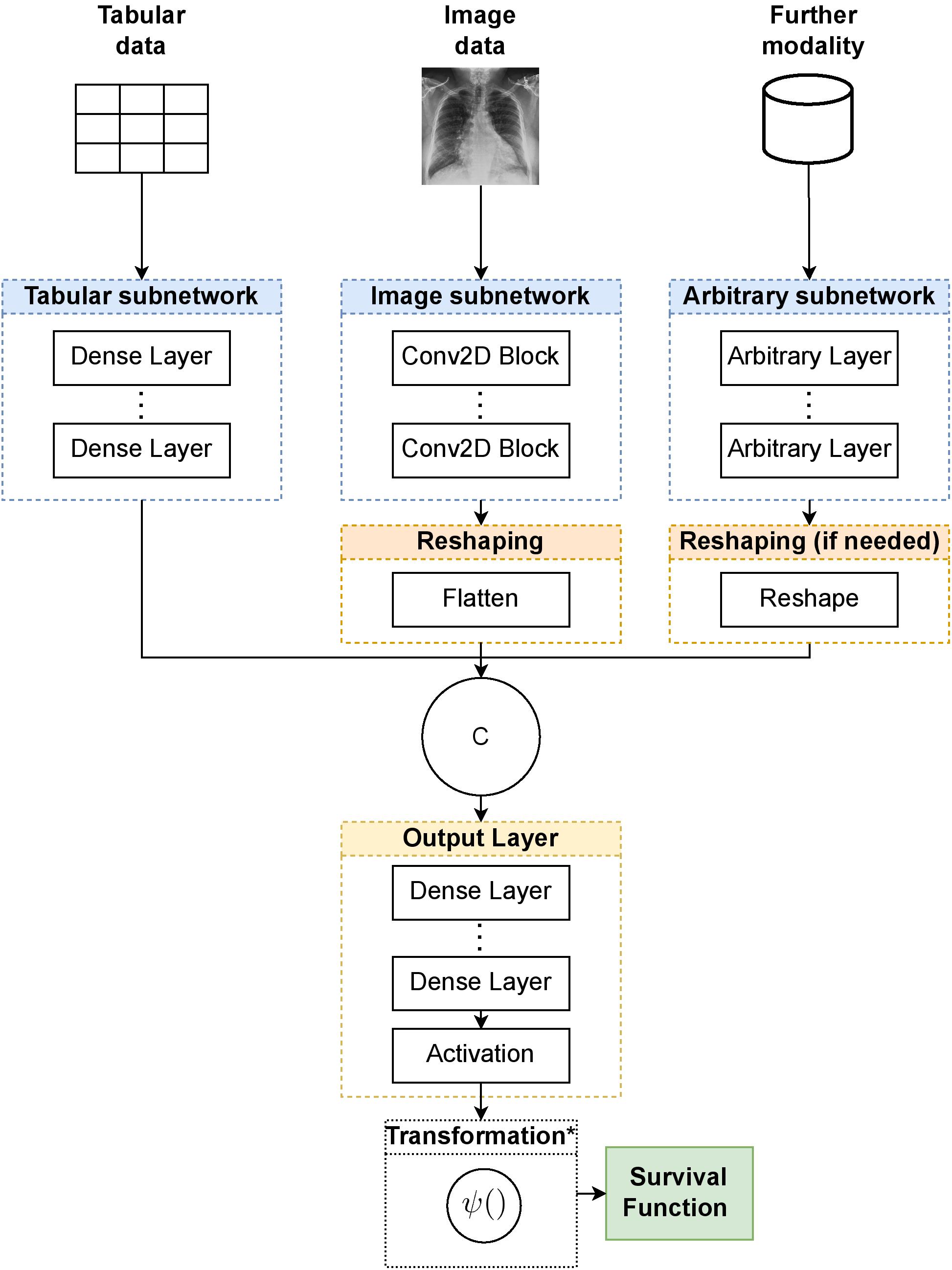}
  \caption{
  Schematic neural architecture for multimodal data input in survival analysis using separate subnetworks for all modalities. Their outputs are reshaped and concatenated to align dimensions. 
  % After passing the flow through the network it is ultimately activated and possibly transformed via $\psi()$ to result in survival probabilities. 
  $\psi()$ transforms the model output to the final outcome.
  The X-ray scan is obtained from \cite{irvin2019chexpert}.
  }
  % \vspace{-0.5cm}
    \label{fig:architecture_MM}
\end{figure*}

Additionally, many methods have shown how to integrate multimodal data, by using a separate subnetwork for each modality.
For instance, one may use a CNN-based subnetwork for image data while tabular data is modeled with an FFNN.
The different modalities can be fused together in different ways in the network head.
If interaction between different modalities is desired, vector representations of the data are concatenated and fed through another joint FFNN.
Otherwise, separate scalars are learned and added onto each other. 
We illustrate two common architectures that tackle competing risks and multiple data modalities --- and can also be combined --- in Figures \ref{fig:architecture_CR} and \ref{fig:architecture_MM}.

\subsection{Estimation and Network Architecture}
\label{Estimation and Network Architecture}

We now review all 61 DL-based survival methods based on theoretical and technical aspects. In Section \ref{Estimation}, we aim to categorize the methods in terms of estimation-related concepts --- model class, loss functions, and parametrization --- and how these concepts correlate. In Section \ref{Network Architecture}, we address the NN architecture choices of all methods reviewed.

\subsubsection{Estimation}
\label{Estimation}

\begin{figure*}[h]
  \centering
    \includegraphics[width=0.8\linewidth]{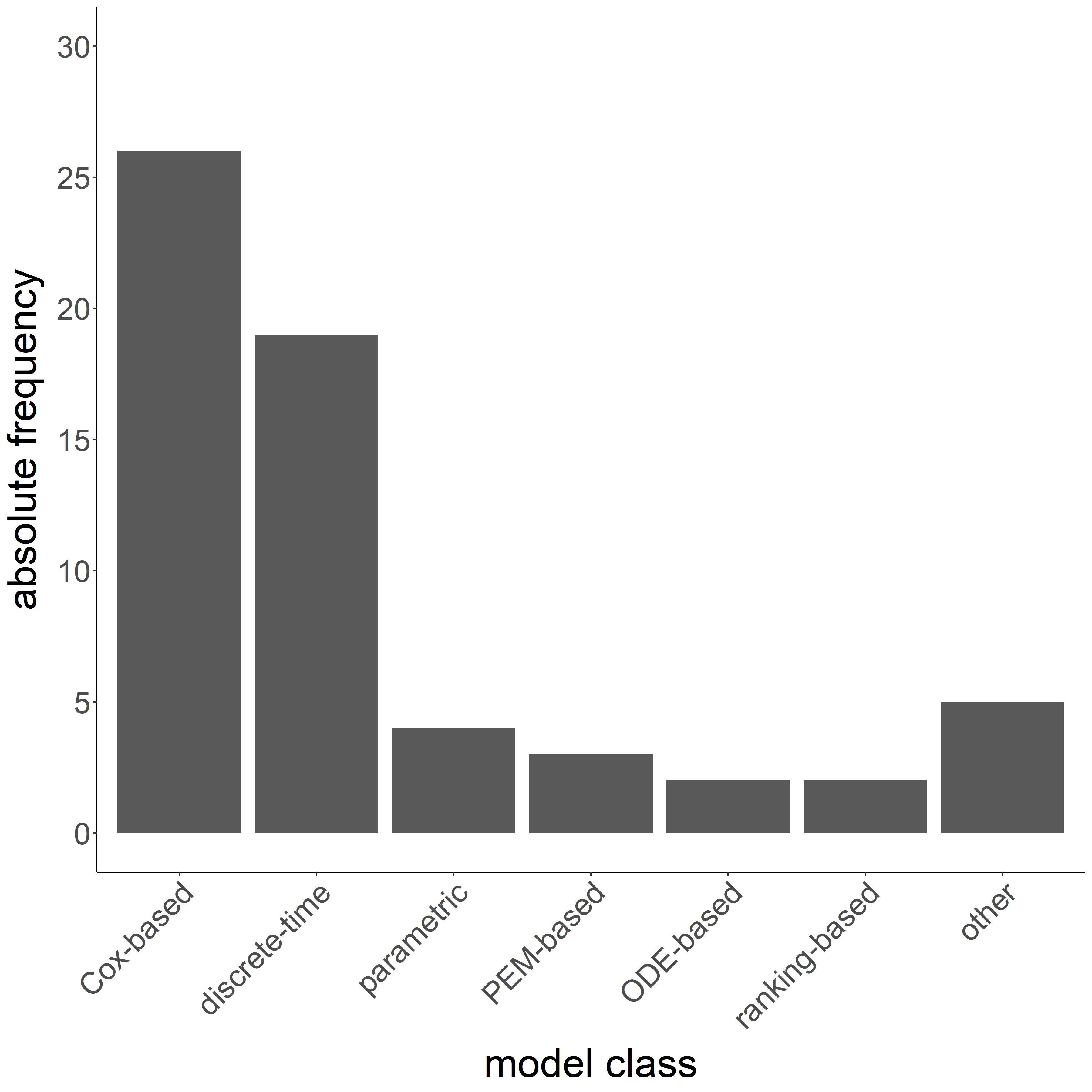}
  \caption{Absolute frequencies of model classes among all 61 methods reviewed.}
  % \vspace{-0.5cm}
    \label{fig:barplots_modelclass}
\end{figure*}

We classify DL-based survival methods in terms of three concepts related to model estimation. First, the \emph{model class} (cf. Figure \ref{fig:barplots_modelclass}) describes which type of statistical survival technique forms the basis of the DL method --- usually one of the approaches introduced in Section \ref{Estimation}. Second, the \textit{loss function} is often a direct consequence of the model class (i.e., its negative log-likelihood). However, as is common in DL, some methods employ multiple losses for improved performance or multi-task learning. For instance, some DL-based survival methods compute a ranking loss, in addition to a standard survival loss, for improvement of the C-index performance measure. The final loss is usually computed as the (weighted) average of all losses applied. Third, the \textit{parametrization} determines which model component is being parametrized by an NN. The standard model parametrization is usually implied by the model class.

Almost all modern DL-based survival methods are optimized with gradient-based methods, featuring tractable loss functions yet with many parameters to be optimized.
Optimizing the loss function in a batch-wise manner, which is the common approach in DL, is not always feasible, though.
This holds for Cox-based methods because the partial loss \eqref{eq:cox_loss} depends on the complete risk set.
Recently, \cite{kvamme_continuous_2019} showed that Cox-based methods can be optimized with stochastic gradient descent methods (i.e., batch-wise) if the batch size is sufficiently large to non-parametrically approximate the risk set.
Before that, deep Cox-based methods were optimized with full gradient descent, making them less attractive for computationally expensive tasks.
% The vast majority of articles analyzed in this review did not focus on optimization techniques for their specific NNs and applied standard optimizers.
% Among those articles that did elaborate on optimization, the Adam optimizer (\cite{2015-kingma}) appears to be the most common choice.

% Figure \ref{fig:barplots_modelclass} depicts the absolute frequencies of model classes among all 61 methods reviewed here.

We now give a detailed description of the above estimation-related concepts as well as their interrelation for all methods reviewed.
% Ideally, all papers should be explicit about these concepts of their methods; unfortunately, this is not always the case.

\paragraph{\textbf{Cox-based Methods}} \mbox{} \\

\vspace{-0.5cm}
\noindent
Out of the 61 methods included in this review, 26 methods are Cox-based; that is, these methods are essentially DL-based modifications and extensions of the Cox regression model. This is underlined by the fact that all of them parametrize the hazard rate --- more precisely, the log-risk function $g(\mathbf{x})$ in \eqref{eq:cox_ph} --- by an NN and minimize the (sometimes slightly modified) Cox loss, i.e., the (negative logarithm of the) partial likelihood of the Cox model.

% As mentioned in Section \ref{history}, the model by \cite{faraggi1995neural} is the first NN-based Cox model, employing only a shallow NN due to computational restrictions at the time. 
\emph{DeepSurv} by \cite{katzman2018deepsurv} extends \cite{faraggi1995neural} by using a deep FFNN as well as different non-linear hidden layer activation functions. The model by \cite{faraggi1995neural} is a simple special case of \emph{DeepSurv}, with a single hidden layer with logistic activation and identity output activation. Note that the PH assumption induced by the Cox PH regression model is maintained in \emph{DeepSurv}, as $g(\mathbf{x})$ remains time-constant despite being parametrized by a (deep) NN. \emph{DeepSurv} uses stochastic gradient descent (SGD) for optimization. To do so, \emph{DeepSurv} uses a restricted risk set including only individuals in the current batch since the Cox loss originally sums over the \textit{entire} risk set, which would impede batching. \emph{Cox-Time} \citep{kvamme2019time} is a more flexible extension of \emph{DeepSurv} where a time-dependent predictor allows estimation of TVEs, i.e. $h(t|\mathbf{x}) = h_0(t)\exp(g(\mathbf{x}, t))$. However, this increased flexibility would render the batching strategy as applied by \emph{DeepSurv} (and most other PH-restricted Cox-based methods) computationally expensive. Therefore, the \emph{Cox-Time} loss function is modified to approximate the risk set by a sufficiently large subset of all individuals at risk, which enables mini-batching and thus scalability to large datasets. \emph{NN-DeepSurv} \citep{tong2022deep} is another extension of \emph{DeepSurv}, employing a nuclear norm for imputation of missing features.

More than half of all Cox-based methods (14) focus on the applicability to high-dimensional data, usually omics data. \emph{MCAP} \citep{chai2022multi} and \emph{VAECox} \citep{kim_improved_2020} both use multiple losses, the latter one within a transfer learning approach. \emph{Cox-nnet} \citep{ching2018cox}, \emph{Cox-PASNet} \citep{hao_cox-pasnet_2018} and its multimodal extension \emph{PAGE-Net} \citep{hao2019page}, \emph{GDP} \citep{xie2019group}, \emph{DNNSurv\_Sun2020} \citep{sun2020genome}, \emph{Qiu2020} \citep{qiu2020meta}, \emph{DeepOmix} \citep{zhao2021deepomix}, and \emph{CNT} \citep{fan2022survival} use simple FFNNs and only a single Cox loss, thus being very similar to \emph{DeepSurv} and \emph{Cox-Time}. \emph{SALMON} \citep{huang2019salmon} and \emph{CNN-Cox} \citep{yin2022convolutional} distinguish themselves through their architecture (see Section \ref{Network Architecture}), \emph{Haa2019} \citep{haarburger_image-based_2019} and \emph{ConcatAE/CrossAE} \citep{tong_deep_2020} through additionally being multimodal (see below).

Eight Cox-based methods focus on unstructured or multimodal input (see also Section \ref{Supported Feature-related Aspects}). \emph{WideAndDeep} \citep{polsterl_wide_2020} combines a linear predictor of tabular features (wide part) with a 1D embedding $d(\mathbf{z})$ learned from a point cloud, which is a latent representation learned from 3D shapes (deep part); subsequently both parts are fused by linearly aggregating the learned weights as in \eqref{eq:predictor_extended}. The model uses the \emph{DeepSurv} loss and thus preserves the PH assumption. \emph{Haa2019} employs a pre-trained CNN of type ResNet18 for subsequent fine-tuning on CT scans, using a Cox loss. Both \emph{DeepConvSurv} \citep{zhu_deep_2016} and \emph{CapSurv} \citep{tang_capsurv_2019} can learn from image data --- yet without incorporating structured (tabular) data --- by using CNN and CapsNet architectures (see Section \ref{Network Architecture}), respectively. \emph{DeepConvSurv} uses a single Cox loss, while \emph{CapSurv} additionally employs the CapsNet margin and reconstruction losses. Both \emph{ConcatAE/CrossAE} and \emph{PAGE-Net} can process high-dimensional data as well as multimodal data; \emph{ConcatAE/CrossAE} use classification and reconstruction losses in addition to the Cox loss to do so, while \emph{PAGE-Net} introduces biologically interpretable pathology, genome-, and a demography-specific layers. \emph{Xie2021} \citep{xie_mixture_2021} can learn from unstructured data for cure rate classification. \emph{DAFT} \citep{wolf2022daft} employs CNNs and a single Cox loss to learn from both structured and unstructured data.

\emph{SurvNet} \citep{wang_survnet_2021} and \emph{DCM} \citep{nagpal2021deep_dcm} do not accommodate any of the additional outcome types or feature-related aspects defined above (see also Section \ref{Supported Survival Tasks}), yet they use multiple losses. In addition to a Cox regression module, \emph{SurvNet} employs an input construction module and a survival classification module (with corresponding losses) for handling missing values and high- versus low-risk profile classification, respectively. \emph{DCM} employs an approximate Monte Carlo Expectation Maximization (EM) algorithm for the estimation of a mixture of Cox models, the total loss also including an Evidence Lower Bound (ELBO) component. \emph{ELMCoxBAR} \citep{wang2019extreme} and \emph{San2020} \citep{sansaengtham2020survival} are standard Cox-based methods in terms of estimation, their architectures being extensions of FFNNs (see Section \ref{Network Architecture}).

\paragraph{\textbf{Discrete-time Methods}} \mbox{} \\

\vspace{-0.5cm}
\noindent
Another 19 methods can be categorized as discrete-time approaches. They consider time to be discrete and usually employ classification techniques, with the outcome being binary event indicators for each discrete time point or interval. The standard loss function of discrete-time DL-based survival methods is the negative log-likelihood (NLL), while typically the discrete hazard \eqref{eq:discrete_hazard} is parametrized by an NN --- just like in the early \emph{PLANN} model. However, as compared to the Cox-based methods which are rather homogeneous methodologically, discrete-time methods are much more heterogeneous in terms of loss functions and architecture.

\emph{DeepHit} \citep{lee2018deephit} is a discrete-time DL-based survival method. It aims to learn first-hitting times directly by not making any assumptions about the underlying stochastic process and parametrizing the discrete PMF directly. \emph{DeepHit} combines two loss functions: first, the log-likelihood derived from the joint distribution of first hitting time and the corresponding event, adjusted for right-censoring and taking into account competing risks; and second, a combination of ranking losses. \emph{Dynamic-DeepHit} \citep{lee2019dynamic}, an RNN-based extension of \emph{DeepHit} which can handle longitudinal input data and thus TVFs, additionally employs a so-called prediction loss for the auxiliary task of step-ahead prediction of TVFs. The transformer-based \emph{TransformerJM} \citep{lin2022deep} also parametrizes the PMF, focusing on modeling survival data and longitudinal data jointly and training on a combination of NLL- and MSE-based losses. 

\emph{RNN-SURV} \citep{giunchiglia_rnn-surv_2018} uses both features and time as inputs, and outputs the survival probability at each discrete time point, employing RNN architecture to use information from previous time points to inform prediction of subsequent time points; the model combines the estimated survival probabilities to a risk score via a weighted sum and employs both an NLL loss (based on the survival probabilities) and a C-index-based (ranking) loss (based on the risk score) for model training.

\emph{Nnet-survival} \citep{gensheimer2019scalable} para\-me\-tri\-zes the discrete hazard \eqref{eq:discrete} by an NN, using an NLL loss as well as mini-batch SGD for rapid convergence and scalability to large datasets. 
Mini-batch SGD is easily applicable to discrete-time methods because the loss only depends on individuals in the current mini-batch.
% --- which is not the case for the Cox loss. This is why \emph{Cox-Time} explicitly modifies its loss function to facilitate batching.) 
% \emph{Nnet-survival} can parametrize the discrete hazard \eqref{eq:discrete} without different types of NNs, though not specifying whether information from more than one data modality can be used. 
The specific architecture --- in particular, the number of neurons per hidden layer and the connectedness of layers --- determines whether TVEs can be modeled or whether the PH restriction is upheld. Another four methods --- \emph{CNN-Survival} \citep{zhang2020cnn}, \emph{MultiSurv} \citep{vale2021multisurv}, \emph{SurvCNN} \citep{kalakoti2021survcnn}, and \emph{Tho2022} \citep{thorsen2022discrete} --- use the same loss and parametrization as \emph{Nnet-survival}. \emph{CNN-Survival} uses a CNN along with transfer learning to learn from CT data (without incorporating tabular data). The multimodal \emph{MultiSurv} first extracts feature representations for each data modality separately, then fuses them, and finally outputs predictions of conditional survival probabilities. \emph{SurvCNN} creates an image representation of multiple omics data types using CNNs and can combine this with clinical data for prediction. \emph{Tho2022} can embed data from multiple modalities, such as electronic health records, and feeds these embedded representations into an RNN which in turn produces survival predictions.

The competing-risk and recurrent-event method \emph{CRESA} \citep{gupta2019cresa} is an RNN-based approach that parametrizes the discrete hazard and uses a loss based on recurrent cumulative incidence functions, which also contains a ranking component as in \emph{DeepHit}. \emph{DRSA} \citep{ren2019deep} also employs an RNN and also parametrizes the discrete hazard, yet as compared to \emph{CRESA} it uses multiple log-likelihood-based losses to predict the likelihood of uncensored events as well as survival rates for censored cases. \emph{Kam2021} \citep{kamran2021estimating} uses the same network architecture as \emph{DRSA}, but proposes a novel training scheme to directly estimate the survival probability: a combination of Rank Probability Score (RPS) loss, emphasizing calibration, and a kernel loss, emphasizing discrimination through penalization of wrongly ordered uncensored individuals. \emph{DCS} \citep{fuhlert2022deep} extends the architecture from \emph{DRSA} and then produces survival probability estimates by employing the same loss function from \emph{Kam2021}, yet modifying the kernel loss component by not only comparing uncensored-uncensored pairs.

\emph{N-MTLR} \citep{fotso_deep_2018} builds upon MTLR and parametrizes the corresponding logistic regression parameters. 
\emph{DNNSurv\_Zha2019} \citep{zhao_dnnsurv_2019} first computes individual-level pseudo (conditional) probabilities, defined as the difference between the estimated survival function with and without individual $i$ and computed on a regular grid of time points, thus reducing the survival task to a regression task, and consequently uses a standard regression loss. \emph{su-DeepBTS} \citep{lee_deepbts_2020} discretizes the time axis but then uses a Cox loss for each time interval, summing up the losses across intervals. \emph{DeepComp} \citep{li_deepcomp_2020} combines distinct losses for censored and uncensored observations with an additional penalty. \emph{SSMTL} \citep{chi_deep_2021} transforms the survival task into a multi-task setting with binary outcome for all time points (or multi-class in case of competing risks), then predicting survival probabilities for each of the time points. \emph{SSMTL} also employs a custom loss made up of a classification loss for uncensored data, a so-called semi-supervised loss for censored data, regularization losses (L1 and L2) as well as a ranking loss in order to ensure monotonicity of predicted survival probabilities.

\emph{Hu2021} \citep{hu_transformer-based_2021}, a transformer-based method, uses an entropy-based loss as well as a discordant-pair penalization loss, parametrizing the discrete hazard. \emph{SurvTRACE} \citep{wang2022survtrace}, another transformer-based method, also parametrizes the discrete hazard, but additionally performs two auxiliary tasks on the survival data: classification and regression; accordingly, the final model loss is a combination of a \emph{PC-Hazard} survival loss (see below), an entropy-based classification loss, as well as a Mean Squared Error (MSE)-based regression loss.

\paragraph{\textbf{Parametric Methods}} \mbox{} \\

\vspace{-0.5cm}
\noindent
The two methods \emph{DeepWeiSurv} \citep{bennis2020estimation} and \emph{DPWTE} \citep{bennis_dpwte_2021} --- the latter one building on the former --- are Weibull-based deep survival methods. Neither of them addresses any of the outcome types or feature-related aspects presented in Section \ref{Data-related Aspects}. \emph{DeepWeiSurv} parametrizes a mixture of Weibull models, as well as both Weibull distribution parameters (see \eqref{eq:aft} with $\theta_1, \theta_2$ the scale and shape parameters of the Weibull distribution), by an FFNN and uses an NLL-based loss function. \emph{DPWTE} employs classification and regression subnetworks to learn an optimal mixture of Weibull distributions, using the same loss function as \emph{DeepWeiSurv} with additional sparsity regularization with respect to the number of mixtures.
\emph{Ava2020} \citep{avati2020countdown} parametrizes the parameters of a log-normal distribution, while being flexible in terms of model architecture. The method introduces the Survival-CRPS loss, a survival adaptation of the Continuous Ranked Probability Score (CRPS). This loss results in well-calibrated survival probabilities and furthermore provides the flexibility to handle both right- and interval-censored data.
\emph{DSM} \citep{nagpal2021_dsm} is a hierarchical generative model based on a finite mixture of parametric primitive distributions similar to the well-known approach by \cite{ranganath2016deep}, using a (mixture) likelihood-based loss as well as an additive loss based on ELBO for uncensored and censored observations; the choice of the parametric survival distribution --- either Weibull or Log-Normal --- is a hyperparameter and can thus be tuned. Its RNN-based extension \emph{RDSM} \citep{nagpal2021_rdsm}, is furthermore capable of handling TVFs.

\paragraph{\textbf{PEM-based Methods}} \mbox{} \\

\vspace{-0.5cm}
\noindent
Three methods rely on the PEM framework to develop a deep survival approach. \emph{PC-Hazard} \citep{kvamme2021continuous} addresses the right-censored single-risk survival task by parametrizing the hazard rate through an FFNN and using the standard likelihood-based PEM loss. Support for other outcome types or feature-related aspects, as introduced in Sections \ref{Outcome Types} and \ref{Feature-related Aspects}, is not discussed. Similarly, \emph{DeepPAMM} \citep{kopper2021semi, kopper2022deeppamm} uses a penalized Poisson NLL as a loss function and also parametrizes the hazard rate by an NN. This method combines a Piecewise Exponential Additive Mixed Model \citep[PAMM;][]{bender2018generalized} with Semi-structured Deep Distributional Regression \citep{ssdr}, which embeds the structured predictor in an NN and further learns from other (unstructured) data types (see \eqref{eq:predictor_extended}).

Finally, \emph{IDNetwork} \citep{cottin2022idnetwork} implements an illness-death model, which uses a PEM-based approach to estimate probabilities for transitions between different states and utilizes FFNNs with shared and transition-specific subnetworks. \emph{IDNetwork} then uses a penalized NLL loss based on the transition probabilities.

\paragraph{\textbf{ODE-based Methods}} \mbox{} \\

\vspace{-0.5cm}
\noindent
\emph{DeepCompete} \citep{aastha_deepcompete_2021} consists of an FFNN shared across all risks as well as an FFNN and a neural ordinary differential equation (ODE) block for each specific risk, using an NLL-based loss. \emph{survNODE} \citep{groha_general_2021} is based on a Markov process and aims to directly solve the Kolmogorov forward equations by using neural ODEs to achieve flexible multi-state survival modeling, with the transition rates parametrized by a nODE architecture (see Sections \ref{Architectural Choices} and \ref{Network Architecture}).

\paragraph{\textbf{Ranking-based Methods}} \mbox{} \\

\vspace{-0.5cm}
\noindent
% For ranking loss-based methods, one can differentiate between methods using ranking losses as auxiliary loss, while for others the ranking loss is the central part of the methodology.
As can be seen in the section above, multiple discrete-time methods (\emph{DeepHit}, \emph{CRESA}, \emph{DCS}, \emph{Kam2021}, \emph{RNN-Surv}, \emph{SSCNN}, and \emph{SSMTL}) use ranking losses as auxiliary losses. 
Beyond that, there are two continuous-time methods --- \emph{RankDeepSurv} and \emph{SSCNN} --- that are built upon ranking losses. Here, we refer to these continuous-time ranking loss-based methods simply as ranking-based methods.
\emph{RankDeepSurv} \citep{jing2019deep} combines ranking losses with an extended MSE loss to augment the number of training samples, without advanced NN architecture or handling of non-standard survival data modalities. 
\emph{SSCNN} \citep{agarwal2021survival} is a multimodal method that reduces histopathology images to whole slide feature maps and uses them, in addition to clinical features, as input of a Siamese Survival CNN; model training with a custom loss --- a combination of a ranking loss with a loss to improve model convergence and pairwise differentiation between survival predictions --- is built directly on the outputs of the Siamese NN.

\paragraph{\textbf{Other Methods}} \mbox{} \\

\vspace{-0.5cm}
\noindent
As for the remaining five methods, \emph{DASA} \citep{nezhad_deep_2019} is a framework introducing a novel sampling strategy based on DL and active learning. The GAN-based \emph{DATE} \citep{chapfuwa2018adversarial} seeks to learn the event time distribution non-parametrically by using adversarial learning and a custom loss function made up of an uncensored-data component, a censored-data component, as well as a distortion loss component. \emph{Hua2018} \citep{huang_deep_2018} uses a CNN architecture and correlational layers for multimodal learning to produce person-specific risks, which are then directly fed into a smooth C-index loss function for model training. \emph{Aus2021} \citep{ausset2021} employs normalizing flows in order to estimate the density of time-to-event data and predict individual survival curves via a transformation model, using an NLL-based loss augmented by an intermediary loss for regularization. Finally, \emph{rcICQRNN} \citep{qin2022survival} is a deep survival method based on a quantile regression NN, parametrizing the quantile regression coefficients by means of an FFNN and using an inverse-probability-of-censoring weighted log-linear quantile regression loss.

% \paragraph{\textbf{Limitations}} \mbox{} \\

% \vspace{-0.5cm}
% \noindent
% Out of all 26 Cox-based methods reviewed here, all but one (\emph{Cox-Time}) assume proportional hazards and thus do not (directly) support TVFs or TVEs; the same applies to all PEM-based methods except \emph{DeepPAMM}. Reliance on the PH assumption --- unless explicit stratification is employed or TVEs are allowed for --- is thus an apparent drawback of Cox- and PEM-based methods. PEM-based approaches furthermore require a pre-transformation of the data, which can slow down estimation depending on how the estimation routine handles it.

% By comparison, discrete-time methods are much more flexible regarding usage of architectures and losses, with many not confined by the PH assumption. However, a disadvantage of assuming time to be non-continuous is that interpolation between the limited, fixed prediction time points is necessary in order to achieve some sort of continuous prediction. 
% Alternatively, one can choose a quasi-continuous grid of discrete event times which, however, may imply much higher computational complexity.

% As for parametric methods, their main disadvantage is of course their distributional assumption --- i.e., the fact that they make an assumption about the distribution of event times.

\subsubsection{Network Architecture}
\label{Network Architecture}

Most DL-based survival methods in this review use FFNNs, often in combination with some other, more advanced architecture. Still, 20 methods --- as well as all early DL-based methods such as the one by \cite{faraggi1995neural} --- exclusively rely on FFNNs. Still, architectural choices among these FFNN-based methods differ. For instance, \emph{DeepHit} uses a softmax outcome layer to produce survival probabilities for each discrete time point and, thus, to model the PMF.

Out of a total of 10 CNN-based methods in this review, eight are multimodal methods that can work with image data: \emph{DeepConvSurv}, \emph{Hua2018}, \emph{Haa2019}, \emph{CNN-Survival}, \emph{PAGE-Net}, \emph{SSCNN}, \emph{Xie2021}, and \emph{DAFT}. 
For instance, \emph{Hua2018} employs CNN and FFNN subnetworks, along with correlational layers, in order to learn from both pathological images and molecular profiles.
The CNN-based method \emph{SurvCNN} is not multimodal per se, but transforms high-dimensional omics data into an image representation in order to feed them into a CNN. \emph{CNN-Cox} combines cascaded Wx \citep{shin2019cascaded}, an NN-based algorithm selecting features based on how well they distinguish between high- and low-risk groups, with a 1D CNN architecture applied to gene expression data. 
Note that the choice of architecture for CNN- and AE-based methods is usually motivated by the objective of extracting information from data input (e.g., from images via CNNs or from omics data via AEs with auxiliary losses), without being very relevant to the target of estimation. This is in contrast to, e.g., RNNs, where the architecture choice is driven by the learning objective.

Nine methods reviewed here use RNN architectures. Six of them --- \emph{RNN-Surv}, \emph{CRESA}, \emph{DRSA}, \emph{Kam2021}, \emph{DCS}, and \emph{Tho2022} --- use a Long Short-Term Memory (LSTM), while the remaining one, \emph{DeepComp}, does not state the RNN architecture it employs. Out of these methods, \emph{RNN-Surv}, \emph{DRSA}, \emph{Kam2021}, and \emph{DCS} do not go beyond the setting of single-risk, right-censored tabular data. For example, \emph{RNN-Surv} uses the RNN to carry forward information from previous time steps, employing a sigmoid output layer activation.
\emph{Tho2022} employs the RNN architecture for multimodal learning from text, medical history, and high-frequency data, while \emph{DeepComp} uses it for competing risk modeling. \emph{CRESA} models both recurrent events and competing risks by means of its RNN architecture. The final two RNN-based methods, \emph{Dynamic-DeepHit} and \emph{RDSM}, are actually extensions of the simpler FFNN-based methods \emph{DeepHit} and \emph{DSM}, respectively, enabling the incorporation of TVFs.

Four methods --- \emph{DASA}, \emph{DCM}, \emph{ConcatAE/CrossAE}, and \emph{VAECox} --- use some form of AEs. 
Another four methods --- \emph{Nnet-survival}, \emph{Ava2020}, \emph{MultiSurv}, and \emph{DeepPAMM} --- do not require a specific architecture, which can instead be flexibly chosen based on application requirements; for instance, a CNN for handling image data (as in \emph{MultiSurv}) or an RNN for incorporating TVFs (as in \emph{Ava2020}). 
Three recent methods, \emph{Hu2021}, \emph{SurvTRACE}, and \emph{TransformerJM}, use a transformer architecture, while another two novel methods, \emph{DeepCompete} and \emph{survNode}, use a nODE architecture. 

Only a single method, \emph{DATE}, uses a GAN architecture (along with a custom loss). \emph{ElmCoxBAR} uses an Extreme Learning Machine (ELM) architecture, which is similar to an FFNN but does not require backpropagation for optimization. \emph{SALMON}, \emph{San2020}, \emph{DPWTE}, and \emph{SurvNet} all use FFNNs, but in a modified manner. \emph{SALMON} adds so-called eigengene modules, using eigengene matrices of gene co-expression modules \citep{zhang2014normalized} instead of raw gene expression data as NN input. \emph{San2020} uses a Stacked Generalization Ensemble Neural Network \citep{wolpert1992stacked}, which takes a combination of \emph{DeepSurv} sub-models and concatenates them for improved hazard prediction. \emph{DPWTE} adds a Sparse Weibull Mixture (SWM) layer to learn the optimal number of Weibull distributions for the mixture model, through an element-wise multiplication of its weights by the previous layer's output. \emph{SurvNet} adds a context-gating mechanism, which is similar to the attention mechanism used in transformers, by adjusting log hazard ratios by survival probabilities from the survival classification module. \emph{WideAndDeep} employs a PointNet \citep{qi2017pointnet} architecture to learn a latent representation of 3D shapes of the human brain while additionally learning from regular tabular data, subsequently fusing both parts. \emph{CapSurv} modifies the CapsNet architecture \citep{sabour2017dynamic}, developed for image classification, by adding a Cox loss and thus making it amenable to SA tasks.

\begin{figure*}[h]
  \centering
    \includegraphics[width=0.8\linewidth]{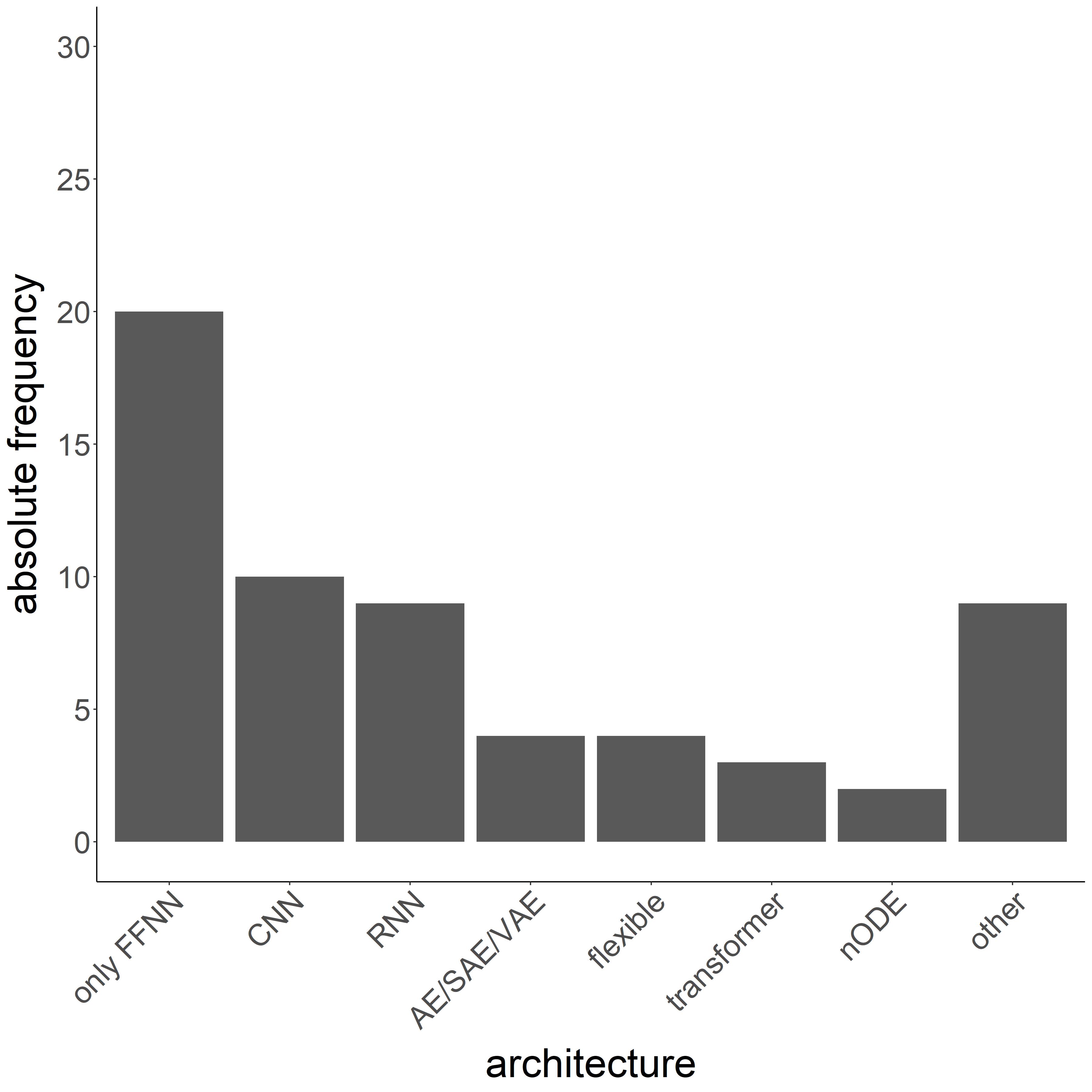}
  \caption{Absolute frequencies of neural network architectures among all 61 methods reviewed.}
    \label{fig:barplots_architecture}
\end{figure*}

Figure \ref{fig:barplots_architecture} depicts the absolute frequencies of NN architectures among all 61 methods included in this review.

\subsection{Supported Survival Tasks}
\label{Supported Survival Tasks}

In this section, we discuss which methods can handle the data-related aspects introduced in Section \ref{Data-related Aspects}. We start by considering \textit{outcome} types and subsequently \textit{feature}-related aspects. Finally, we summarize which methods offer (which kind of) interpretability of results.
% The majority of DL-based survival methods in this review focus on the most classical survival scenario: tabular time-constant features and single-risk right-censored outcomes.

\subsubsection{Supported Outcome Types}
\label{Supported Outcome Types}

Regarding censoring and truncation of event times, left-censoring and right-truncation are not explicitly addressed by any of the methods reviewed.
\emph{Ava2020} is capable of handling interval-censored data thanks to the flexibility of the Survival-CRPS loss.
\emph{survNode} briefly addresses interval-censoring and left-truncation by stating how they would affect likelihood computations. 
\emph{DSM} mentions that the modeling framework is amenable to these two output modalities. 
In \emph{DeepPAMM} left-truncation is accounted for in the data pre-processing step.

\begin{figure*}[hb]
  \centering
    \vspace{-0.4cm}
    \includegraphics[width=0.99\linewidth]{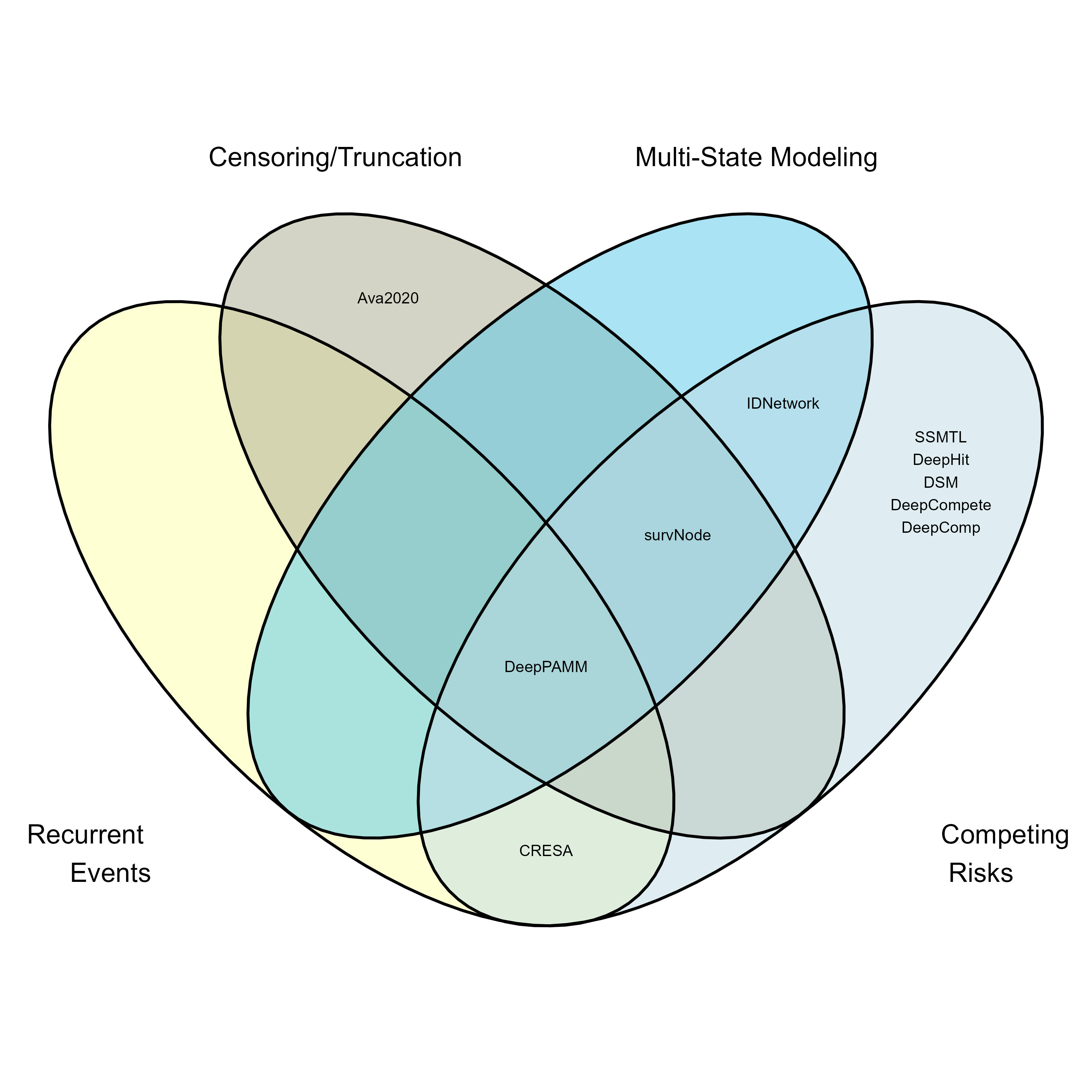}
    \vspace{-1.5cm}
  \caption{Venn diagram illustrating which methods can handle the distinct survival outcome types.}
  \label{fig:venn_outcomes}
  % \vspace{-0.5cm}
\end{figure*}

Nine methods are designed to deal with competing risks; interestingly, none of these methods is Cox-based, and four of them are discrete-time. \emph{DeepHit}, \emph{CRESA}, and \emph{DeepComp} all assume time to be discrete and employ cause-specific subnetworks, with \emph{DeepHit} using FFNNs to generate a final distribution over all competing causes for each individual; both \emph{CRESA} and \emph{DeepComp} use RNN architectures, yet while \emph{CRESA} also generates a final distribution over all competing causes, \emph{DeepComp} outputs cause-specific discrete hazards for each time interval. \emph{SSMTL}, also discrete-time, uses an FFNN architecture, views competing risk SA as a multiclass problem, and creates a custom loss with separate components for non-censored and censored individuals, as well as a ranking component. \emph{DeepCompete} is a continuous-time method that employs nODE blocks within each of its cause-specific subnetworks in order to output a cumulative hazard function. \emph{DSM} first learns a common representation of all competing risks by passing through a single FFNN. Based on this representation, and treating all other events as censoring, the event distribution for a single risk is then learned using cause-specific Maximum Likelihood Estimation (MLE); the ELBO loss is also adjusted to treat competing events as censoring. Both \emph{survNode} and \emph{IDNetwork} are based on Markov processes --- illness-death process and Markov jump process, respectively --- and thus naturally handle competing risks and even the more general case of multi-state outcomes. Being PEM-based, \emph{DeepPAMM} parametrizes the hazard rate, which is a transition rate by definition; \emph{DeepPAMM} can further specify multiple transitions and therefore model competing risks as well as multi-state outcomes. Finally, two methods discuss handling of recurrent events: \emph{CRESA} employs an RNN architecture with time steps representing recurrent events, while \emph{DeepPAMM} uses random effects inspired by statistical mixed models. Figure \ref{fig:venn_outcomes} summarizes which outcome types beyond right-censoring the methods reviewed explicitly mention.

\subsubsection{Supported Feature-related Aspects}
\label{Supported Feature-related Aspects}

One important feature-related aspect is time dependence, a deviation from the PH assumption imposed by traditional survival models such as Cox regression or Weibull AFT. Seven methods address TVFs: \emph{DeepHit}'s and \emph{DSM}'s RNN-based extensions \emph{Dynamic-DeepHit} and \emph{RDSM}, as well as \emph{CRESA}, \emph{Ava2020} (by choosing an RNN architecture), \emph{survNode}, \emph{DeepPAMM}, and \emph{TransformerJM}. The technical incorporation of TVFs is, for example, achieved by converting tabular time-varying feature input into long format (\emph{DeepPAMM}) or by employing RNNs prior to each new feature measurement (\emph{survNode}).

TVEs constitute another deviation from the PH assumption: Seven methods are capable of modeling effects that might not be constant over time, with four of them being time-discrete approaches. \emph{Nnet-survival} and \emph{MultiSurv} incorporate TVEs modeling by using a fully connected NN to connect the final hidden layer's neurons with the output nodes, while \emph{RNN-Surv} captures TVEs through its RNN architecture. \emph{Cox-Time} accommodates TVEs by making the Cox-style relative risk --- which it parametrizes by an NN --- time-dependent and \emph{DeepPAMM} can address TVEs through the interaction of the follow-up time (represented as a feature) with other features.
\emph{DSM} and \emph{SSMTL} do not provide further detail about how TVEs are being estimated.

Another feature-related aspect is the integrability of high-dimensional (usually omics) data, which implies learning from a high-dimensional predictor space. While all DL-based methods are generally capable of handling high-dimensional feature inputs, here we focus on the 18 DL-based survival methods that are explicitly designed to work with high-dimensional data, usually by applying specialized regularization techniques. 14 of these methods --- \emph{Cox-nnet}, \emph{Cox-PASNet} and \emph{PAGE-Net}, \emph{Haa2019}, \emph{GDP}, \emph{SALMON}, \emph{ConcatAE/CrossAE}, \emph{DNNSurv\_Sun2020}, \emph{Qiu2020}, \emph{VAECox}, \emph{DeepOmix}, \emph{CNN-Cox}, \emph{CNT}, and \emph{MCAP} --- are (partially) Cox-based. As for the remaining four methods, \emph{CNN-Survival}, \emph{MultiSurv}, and \emph{SurvCNN} are discrete-time methods, while \emph{rcICQRNN} is quantile regression-based.

Finally, a total of 16 methods can (hypothetically) extract information from unstructured or multimodal features. Eight of them are (partially) CNN-based, underlining the focus on processing mostly medical image data. \emph{DeepConvSurv}, \emph{CapSurv}, and \emph{CNN-Survival} (the last one employing transfer learning) exclusively work with imaging data without incorporating any tabular information, which is why these methods are not truly \textit{multi}modal. Similarly, \emph{Nnet-survival}, being flexible in terms of NN architecture, can learn from image data by choosing a CNN, yet again at the cost of discarding tabular data as only a single data modality can be handled. \emph{Hua2018} incorporates both image and molecular data yet without making any mention of tabular data. \emph{Haa2019} fine-tunes a pre-trained ResNet18, optionally concatenating it with radiomics features, and additionally leverages clinical data. 

\emph{PAGE-Net} employs a novel patch aggregation strategy to integrate unstructured Whole Slide Images (WSIs) and structured demographic and genomic data. \emph{SSCNN} creates feature maps from WSIs and employs a Siamese CNN to learn from both these feature maps as well as clinical features. \citep{liu2022deep} also use DL to extract features from WSIs in the context of survival analysis, however, not in an end-to-end approach within the network. 

\begin{figure*}[h]
  \centering
    \vspace{-0.5cm}
    \includegraphics[width=0.99\linewidth]{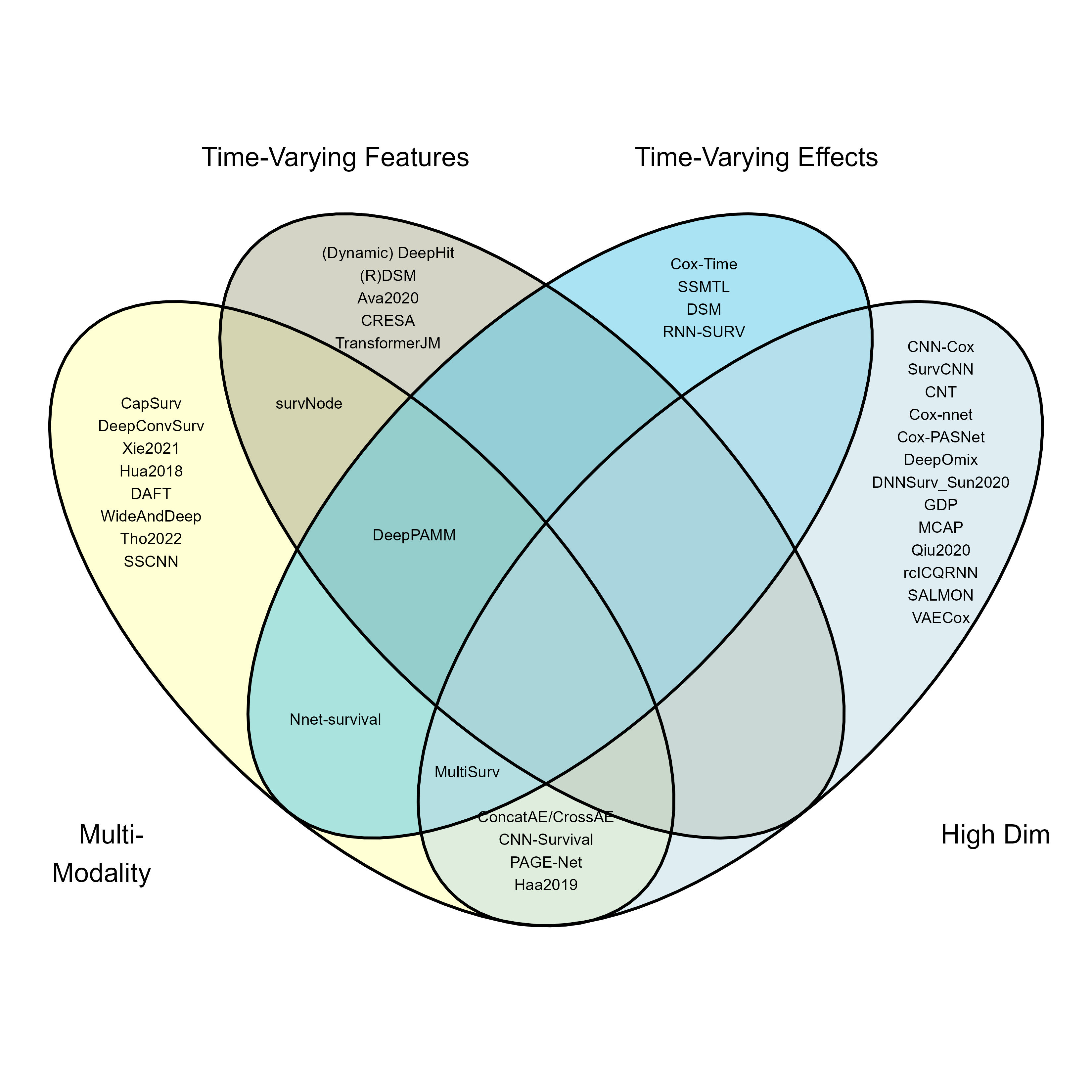}
    \vspace{-1.5cm}
  \caption{Venn diagram illustrating which methods can handle the distinct survival feature-related aspects.}
  \label{fig:venn_features}
  \vspace{-0.5cm}
\end{figure*}

\emph{ConcatAE/CrossAE} integrates information from multiple modalities, either through modality-specific autoencoders or cross-modality translation; the integration of tabular data is, however, not explicitly mentioned.
\emph{survNode} can conceptually account for multimodal features by encoding initial values with, e.g., CNN or NLP layers.
The cure rate model \emph{Xie2021} only allows for (single-modality) unstructured data for determining the cure rate probability through a CNN.
\emph{DAFT} uses a ResNet CNN architecture as its backbone, feeding tabular data into it through a novel Dynamic Affine Feature Map Transform (DAFT) module, which in turn enables a bidirectional information flow between image and tabular data. Finally, \emph{Tho2022} employs an RNN architecture to create an embedding for electronic patient record data (such as medical history and free text) and further fuses tabular clinical features into the model before generating survival predictions. \emph{WideAndDeep}, using a Alzheimer's Disease (AD) dataset, learns a latent representation of 3D shapes of the human brain while additionally learning from regular tabular data, subsequently fusing both parts. \emph{MultiSurv}, a multimodal extension of \emph{Nnet-survival}, and \emph{DeepPAMM} both provide flexibility in terms of architecture choice so that, for example, image data could be incorporated by employing CNNs for the NN part; they also fuse information from the different data modalities. 

Figure \ref{fig:venn_features} illustrates which of the methods incorporate the different types of feature-related aspects.

\subsection{Interpretability}
\label{Interpretability}

\noindent
% The high complexity of DL (and ML) methods usually means that they do not provide the same degree of interpretability as the survival models considered in Section \ref{sec:Estimation}. 
% Yet in fields like the life sciences, results must be interpretable to provide a solid basis for highly sensitive decision-making \citep{vellido2020importance}. 
% Our \textit{Main Table} summarizes which of the methods provide interpretable results. 
% Some methods offer inherently interpretable results --- e.g., classical statistical interpretability of tabular features --- or assign biological meaning to the NN nodes. Other methods use post-hoc interpretability methods such as \emph{LIME} \citep{ribeiro2016should} or \emph{SHAP} \citep{lundberg2017unified}. Furthermore, survival-adapted extensions of (model-agnostic, post-hoc) interpretability techniques such as \emph{SurvNAM} \citep{utkin2022survnam} or \emph{SurvSHAP(t)} \citep{krzyzinski2022survshap} exist.

By construction, DL methods (as well as ML methods) are more complex than the survival models considered in Section \ref{sec:Estimation} and thus usually do not provide the same degree of interpretability. At the same time, in fields such as the life sciences, results and model outputs must be interpretable to provide a solid basis for highly sensitive decision-making \citep{vellido2020importance}. Here, we summarize which of the methods provide (inherently) interpretable results.

\emph{Cox-nnet}, \emph{Cox-PASNet}, \emph{PAGE-Net} and \emph{DeepOmix} provide some interpretability by assigning biological meaning to the nodes of their NNs. 
\emph{Cox-nnet} obtains biologically relevant hidden nodes, as the most variable nodes can be viewed as surrogate features for discriminating patient survival and, in addition, these nodes correlate strongly with significantly enriched pathways.
\emph{Cox-PASNet} and \emph{PAGE-Net} both possess genome-specific layers, which include a gene input layer, a pathway layer embedding prior pathway-related knowledge for biological interpretation, and data integration layers. 
The two methods then rank the node values of features by the average absolute partial derivatives (with respect to the data integration layers) for a pathway-based interpretation of genomic data: explicitly, pathway nodes each represent a biological pathway. 
\emph{PAGE-Net} additionally possesses pathology-specific layers, which identify features relevant to SA from histopathological images via pre-trained CNNs; at the patient-level, these survival-discriminatory features eventually represent a histopathological WSI. 
\emph{DeepOmix} incorporates multi-omics data via a gene input layer and prior biological and pathway knowledge via functional module layers (low-dimensional representations), guided by the idea that genes do not work in isolation but rather function as functional modules. 
With each node representing a non-linear function of the genes' different attributes (e.g., mutations), \emph{DeepOmix} obtains biological interpretability because it captures the (non-linear) effects of biological pathways onto survival time.

By fusing the output of an NN for image data with the output of a Cox PH model for tabular data, \emph{WideAndDeep} retains the interpretability of a standard Cox regression for structured features. 
\emph{Xie2021} also provides standard Cox model interpretability, because survival prediction is performed through non-deep Cox regression. 
\emph{DeepPAMM} provides classical statistical interpretability of the structured effects by its architecture, with identifiability, if necessary, ensured through orthogonalization \citep{rugamer.new.2023}. \emph{survNode} introduces a latent variable extension providing aspects of feature interpretability. 
The transformer-based \emph{SurvTrace} method makes use of attention maps, comparing attention scores of different features across selected individuals to provide some interpretability of feature effects.

It is worth noting that post-hoc methods from the field of Interpretable Machine Learning and explainable AI, such as Permutation Feature Importance \citep{breiman2001random}, Local Interpretable Model-agnostic Explanations \citep[LIME][]{ribeiro2016should}, Shapley Additive exPlanations \citep[SHAP][]{lundberg2017unified}, attention maps \citep{jetley2018learn}, Layer-Wise Relevance Propagation \citep[LRP][]{montavon2019layer}, and Neural Additive Models \citep[NAMs;][]{agarwal2021neural}, are potentially applicable to DL-based survival methods. 
However, this is subject to current research and it is not always clear if and how such methods need to be adjusted to account for different types of censoring, truncation or other outcome types. 

Several survival-specific adaptations of the abovementioned post-hoc interpretability methods have already been developed; for instance, \emph{SurvLIME} \citep{kovalev_survlime_2020}, \emph{SurvSHAP(t)} \citep{krzyzinski2022survshap}, and \emph{SurvNAM} \citep{utkin2022survnam} are based on \emph{LIME}, \emph{SHAP}, and \emph{NAMs}, respectively. \cite{cho2023interpretable} use meta-learning and the \emph{DeepLIFT} \citep{shrikumar2017learning} method to make the integration of multi-omics data in SA more interpretable.

Among the papers reviewed here, \emph{SALMON} explores feature importance of individual inputs, \emph{DNNSurv\_Sun2020} employs LIME, \emph{Tho2022} uses SHAP, and \emph{SSMTL} computes post-hoc feature importance and plots feature effects on cumulative incidence curves. \emph{Qiu2020} uses a risk propagation technique called \emph{SurvivalNet} \citep{yousefi2017predicting}, which is an explanation method specific to SA.

\subsection{Model Evaluation and Comparison}
\label{Model Comparison and Evaluation}

Model evaluation is an important aspect of any machine learning pipeline, and SA in particular.
Typical metrics in benchmark experiments of survival models are the C-index (usually Harrell's \citep{harrell1982} or Uno's \citep{uno2007}) for assessing risk predictions, and the Brier/Graf score \citep{graf1999} for evaluation of distribution predictions, with the C-index being by far the most popular metric among the methods reviewed here. Typically underreported are the right-censored logloss \citep{avati2020countdown} and calibration measures such as D-calibration \citep{haider2020effective}. Recent work also suggests that most of the previously used evaluation measures in SA do not constitute proper scoring rules \citep{sonabend2022scoring}. Proper alternatives have been proposed recently \citep{rindt2022, sonabend2022scoring}, but have not been widely adopted yet. 

Interpretation and comparison of the self-reported benchmark experiments in different articles is often not meaningful for various reasons: 
The datasets used, their pre-processing, and handling of missing values is not the same. 
Even if the same data sets are used, the definition of resampling strategy, the exact definition of the respective evaluation metrics (e.g. different variants of C-Index, integration window of the integrated brier score, etc.) and their use (e.g. transformation of survival distribution predictions for measures of discrimination \citep{sonabend.avoiding.2022}) are often not clearly specified or not identical.
Further general issues that hinder direct interpretation or reported results are potential issues of selective reporting and researchers degrees of freedom (selection of data sets, choice of evaluation metrics, decisions about budget and hyperparameter space for tuning of the proposed as well as competing algorithms, etc.) that have plagued applied sciences but have also been bemoaned in methodological research \citep[e.g.][]{boulesteix.replication.2020, niessl.overoptimism.2022}. 

For all these reasons, direct comparison of the performance of different methods reviewed in this article is not possible. 
This calls for future research to conduct neutral benchmark studies \citep{boulesteix.plea.2013}. Such an investigation has been for example conducted for some non-DL-based ML methods on omics data \citep{herrmann2021large}. 
However, such studies are generally hard to conduct and require substantial effort, in particular for DL-based methods with high computational requirements, and because general purpose implementations of most of the methods reviewed here are not available and code repositories are missing for almost half of the methods (cf. Section \ref{Reproducibility}). 

\subsection{Sample Size Requirements}
\label{Sample Size Requirements}
Sample size considerations are an equally important topic that needs further research in the context of DL-based survival analysis.

In general, the sample size required for training a DL-based method crucially depends on the model architecture (such as the choice of network architecture and hyperparameters or the use of transfer learning) as well as on the input data modalities (e.g., whether images or high-dimensional omics data are being used), and the assumed data generating process. 
In addition, sample size calculations are very task-specific: \cite{fang2021impact} show that the required sample size for organ auto-segmentation critically depends on the organ to be segmented. 
Overall, sample size calculation in DL is still quite rare, being an active field of research itself \citep{shahinfar2020many, fang2021impact}. For instance, in ML-based medical imaging analysis, a systematic review of methodologies for sample size calculation by \cite{balki2019sample} identified only four such methods, highlighting the need for future work in this area.

This is particularly true for DL-based SA, as, to our knowledge, there is currently no research published on sample size calculation in this specific area. Generally, in SA, the power for detection of effects does not depend on the overall sample size but rather on the number of events (for a specific transition). As a consequence, censoring, truncation and other outcome-related specifics need to be taken into account. For example, effects on the development of a rare condition could be hard to detect if there is a competing event with high prevalence. Additionally taking into account imaging data will generally make the assumed data generating process and therefore sample size calculation more complex.
As for more complex statistical models, simulation-based sample size calculation could be a way to go in the future \citep{snell.external.2021}. 

The papers reviewed in this work do not explicitly address sample size requirements.
In our \textit{Main Table} we included a column that indicates the minimum dataset size among all benchmarked datasets used for each method. However, this answers a different question about applicability. Most of the methods reviewed will be applicable to rather small data sets, however, their ability to learn anything and outperform simpler baseline models will usually decrease with diminishing sample size. 

\subsection{Reproducibility}
\label{Reproducibility}

\noindent
Code and data accessibility foster open and reproducible research. 
The availability of code can indicate a method's maturity and its general applicability to new use cases. 
However, the code of algorithms and benchmark experiments is not publicly accessible for 25 methods. 
Furthermore, the accompanying codes of 28 methods are one-shot implementations and have not yet been processed into easy-to-use packages.
Data availability ensures that the reported results can be reproduced and are available for future benchmark experiments. 
% All papers reviewed feature at least one accessible data set per exclusion criteria.
% All papers reviewed in this work use accessible (i.e., public or registered access) benchmark data by construction (see exclusion criteria), with many of them benchmarking their method on multiple datasets as well as on synthetic data.
The \textit{Main Table} summarizes reproducibility aspects (in terms of code and data) for all methods.

For usability and reproducibility, new methods should ideally be packaged and also integrated within one of the general purpose suits for machine learning and benchmarking for survival analysis such as \texttt{auton-survival} \citep{nagpal2022auton}, \texttt{mlr3proba} \citep{mlr3proba}, \texttt{pycox} \citep{kvamme2019time}, \texttt{scikit-survival} \citep{sksurv}, or similar.

% \subsection{Further Challenges}
% \label{Challenges}

% For applications, we regard hyperparameter tuning as particularly crucial in deep survival analysis.
% Some of the contributions listed in this review explicitly discuss hyperparameter tuning, but it remains globally underaddressed.
% Particularly, regularization is critical given the fact that censoring usually takes away a lot of information from survival data sets.
% The need for further tuning and validation data will shrink the data set even further making it harder for the methods to learn complex associations.
% This is especially true when unstructured data, such as images, are employed.

\section{Conclusion}
\label{Conclusion}

SA is concerned with modeling the time until an event of interest occurs while accounting for censoring, truncation, and other aspects of time-to-event data (cf. Section \ref{Data-related Aspects}). 

In this paper, we provide a structured, comprehensive review of DL-based survival methods, from a theoretical as well as practical perspective. In doing so, we aim to enable practitioners to quickly gauge the methods available for their specific use case as well as to help researchers to identify the most promising areas for future research. The main results are summarized in an open-source, interactive, editable table (\href{https://survival-org.github.io/DL4Survival}{https://survival-org.github.io/DL4Survival}). All data, figures, and code scripts used in this work can be found in the corresponding repository (\href{https://github.com/survival-org/DL4Survival}{https://github.com/survival-org/DL4Survival}).

We conclude that most methodologically innovative DL-based survival methods are survival-specific applications of novel methods developed in other areas of DL, such as computer vision or NLP. This usually yields a more flexible estimation of associations of (structured and unstructured) features with the outcome, rather than solving problems of time-to-event data not addressed by, e.g., statistical approaches. Outcome types beyond right-censoring and competing risks are rarely addressed, potentially due to a limited number of application cases.

Further, little attention has been paid to optimization (e.g., choice of optimizers, tuning of hyperparameters, or neural architecture search) among the methods reviewed here, as they usually focus on network architecture, data modalities, and specific use cases. Among those articles that did elaborate on optimization, the Adam optimizer \citep{2015-kingma} appears to be the most common choice.

There are also some challenges specific to DL-based SA. In the parametric setting, many common log-likelihood-based losses for survival analysis are poorly conditioned. 
For example, modeling a Weibull distribution that assumes errors from an extreme value distribution (with standardized density $f(t) = \exp(-t)\exp(-\exp(-t))$) may be particularly challenging when being optimized with gradient descent and low precision. 
Similarly, \cite{avati2020countdown} recommend the log-normal distribution since optimization of other distributions that are suitable for time-to-event data suffers from numerical instability, as their densities have forms of type $(t\theta_1)^{\theta_2}$ (where $\theta_1$ and $\theta_2$ are parameters of interest) or contain the Gamma function. Their optimization will be particularly challenging, when all parameters of a distribution are learned depending on features (cf. \eqref{eq:aft}).
Batching is another issue specific to DL-based optimization. 
In semi-parametric models like the Cox model, batching might become problematic, as already discussed in \ref{Estimation}. 
More generally, batching might need to be adapted, depending on the survival task. In recurrent events settings, for example, batching might need to be set up differently, depending on whether one wants to predict next recurrence for all subjects (given previous recurrences) or the entire process for a new subject. 
Finally, the lack of openly accessible, high-dimensional, potentially multimodal datasets remains a major challenge to the development and training of novel DL-based survival methods.

Missing values are rarely discussed within the methods we reviewed; indeed, most methods implicitly require missing values to be taken care of during data preprocessing. Explicit handling of missing values in the time-to-event setting is done only by \emph{MultiSurv} and \emph{SurvNet}.

In terms of their application, DL-based survival methods have been deployed in estimating patient survival based on medical images (usually CT scans of a particular anomaly) or (multi-)omics data --- as evidenced by the large majority of multimodal or high-dimensional methods in this review. Moreover, some methods are explicitly motivated by a specific clinical use case: \emph{DASA} by prostate cancer; \emph{Haa2019}, \emph{su-DeepBTS}, and \emph{SurvNet} by lung cancer; \emph{SALMON}, \emph{ConcatAE/CrossAE}, and \emph{Liu2022} by breast cancer; and \emph{MCAP} by ovarian cancer. 
Other areas of application of DL-based survival methods include improved estimation of prognostic indices \citep{bice2020deep} and of recurrence after cancer surgery \citep{lee_deepbts_2020}. The choice of datasets used for benchmarking (see \textit{Main Table}) provides further information about the application cases for each method. 

In summary, deep survival methodology has advanced substantially in recent years and will certainly continue to benefit from developments in ML/DL, with big methodological advances being likely to swap over. 
In particular, generative DL techniques like diffusion are promising candidates for adaptation to survival tasks. 
% The lack of openly accessible, high-dimensional, potentially multimodal datasets remains a major challenge to the development and training of novel DL-based survival methods. 
The rapid progress in this area of research is also why any overview work can never be fully exhaustive or up-to-date. Therefore, we actively encourage the research community to contribute to our open-source interactive table (\href{https://survival-org.github.io/DL4Survival}{https://survival-org.github.io/DL4Survival}).

% SA will certainly continue to benefit from developments in ML/DL, with big methodological advances being likely to swap over. In particular, generative DL techniques like diffusion are promising candidates for adaptation to survival tasks. However, a bottleneck regarding the application of advanced DL methods to SA is the lack of openly accessible, high-dimensional, potentially multimodal datasets.

% \bibliography{reference}
\clearpage
\printbibliography

@article{agarwal2021neural,
  title={Neural additive models: Interpretable machine learning with neural nets},
  author={Agarwal, Rishabh and Melnick, Levi and Frosst, Nicholas and Zhang, Xuezhou and Lengerich, Ben and Caruana, Rich and Hinton, Geoffrey E},
  journal={Advances in neural information processing systems},
  volume={34},
  pages={4699--4711},
  year={2021}
}

@inproceedings{agarwal2021survival,
  title={Survival prediction based on histopathology imaging and clinical data: A novel, whole slide cnn approach},
  author={Agarwal, Saloni and Eltigani Osman Abaker, Mohamedelfatih and Daescu, Ovidiu},
  booktitle={Medical Image Computing and Computer Assisted Intervention--MICCAI 2021: 24th International Conference, Strasbourg, France, September 27--October 1, 2021, Proceedings, Part V 24},
  pages={762--771},
  year={2021},
  organization={Springer}
}

@INPROCEEDINGS{ausset2021,
  author={Ausset, Guillaume and Ciffreo, Tom and Portier, Francois and Clémençon, Stephan and Papin, Timothée},
  booktitle={2021 IEEE 8th International Conference on Data Science and Advanced Analytics (DSAA)}, 
  title={Individual Survival Curves with Conditional Normalizing Flows}, 
  year={2021},
  volume={},
  number={},
  pages={1-10},
  doi={10.1109/DSAA53316.2021.9564222}}

@inproceedings{avati2020countdown,
  title={Countdown regression: sharp and calibrated survival predictions},
  author={Avati, Anand and Duan, Tony and Zhou, Sharon and Jung, Kenneth and Shah, Nigam H and Ng, Andrew Y},
  booktitle={Uncertainty in Artificial Intelligence},
  pages={145--155},
  year={2020},
  organization={PMLR}
}

@inproceedings{ballard1987modular,
  title={Modular learning in neural networks.},
  author={Ballard, Dana H},
  booktitle={Aaai},
  volume={647},
  pages={279--284},
  year={1987}
}

@article{bender2018generalized,
  title={A generalized additive model approach to time-to-event analysis},
  author={Bender, Andreas and Groll, Andreas and Scheipl, Fabian},
  journal={Statistical Modelling},
  volume={18},
  number={3-4},
  pages={299--321},
  year={2018},
  publisher={Sage Publications Sage India: New Delhi, India}
}

@inproceedings{bennis2020estimation,
  title={Estimation of conditional mixture Weibull distribution with right censored data using neural network for time-to-event analysis},
  author={Bennis, Achraf and Mouysset, Sandrine and Serrurier, Mathieu},
  booktitle={Advances in Knowledge Discovery and Data Mining: 24th Pacific-Asia Conference, PAKDD 2020, Singapore, May 11--14, 2020, Proceedings, Part I 24},
  pages={687--698},
  year={2020},
  organization={Springer}
}

@article{biganzoli1998feed,
  title={Feed forward neural networks for the analysis of censored survival data: a partial logistic regression approach},
  author={Biganzoli, Elia and Boracchi, Patrizia and Mariani, Luigi and Marubini, Ettore},
  journal={Statistics in medicine},
  volume={17},
  number={10},
  pages={1169--1186},
  year={1998},
  publisher={Wiley Online Library}
}

@article{binder2008allowing,
  title={Allowing for mandatory covariates in boosting estimation of sparse high-dimensional survival models},
  author={Binder, Harald and Schumacher, Martin},
  journal={BMC bioinformatics},
  volume={9},
  number={1},
  pages={1--10},
  year={2008},
  publisher={Springer}
}

@article{breiman2001random,
  title={Random forests},
  author={Breiman, Leo},
  journal={Machine learning},
  volume={45},
  pages={5--32},
  year={2001},
  publisher={Springer}
}

@article{brown1997use,
  title={On the use of artificial neural networks for the analysis of survival data},
  author={Brown, Stephen F and Branford, Alan J and Moran, William},
  journal={IEEE transactions on neural networks},
  volume={8},
  number={5},
  pages={1071--1077},
  year={1997},
  publisher={IEEE}
}

@inproceedings{chai2022multi,
  title={A Multi-constraint Deep Semi-supervised Learning Method for Ovarian Cancer Prognosis Prediction},
  author={Chai, Hua and Guo, Longyi and He, Minfan and Zhang, Zhongyue and Yang, Yuedong},
  booktitle={Advances in Swarm Intelligence: 13th International Conference, ICSI 2022, Xi'an, China, July 15--19, 2022, Proceedings, Part II},
  pages={219--229},
  year={2022},
  organization={Springer}
}

@inproceedings{chapfuwa2018adversarial,
  title={Adversarial time-to-event modeling},
  author={Chapfuwa, Paidamoyo and Tao, Chenyang and Li, Chunyuan and Page, Courtney and Goldstein, Benjamin and Duke, Lawrence Carin and Henao, Ricardo},
  booktitle={International Conference on Machine Learning},
  pages={735--744},
  year={2018},
  organization={PMLR}
}

@inproceedings{chen2018neural_2,
author = {Chen, Ricky T. Q. and Rubanova, Yulia and Bettencourt, Jesse and Duvenaud, David},
title = {Neural ordinary differential equations},
year = {2018},
publisher = {Curran Associates Inc.},
address = {Red Hook, NY, USA},
booktitle = {Proceedings of the 32nd International Conference on Neural Information Processing Systems},
pages = {6572–6583},
numpages = {12},
location = {Montr\'{e}al, Canada},
series = {NIPS'18}
}

@article{ching2018cox,
  title={Cox-nnet: an artificial neural network method for prognosis prediction of high-throughput omics data},
  author={Ching, Travers and Zhu, Xun and Garmire, Lana X},
  journal={PLoS computational biology},
  volume={14},
  number={4},
  pages={e1006076},
  year={2018},
  publisher={Public Library of Science San Francisco, CA USA}
}

@article{cho2023interpretable,
  title={Interpretable meta-learning of multi-omics data for survival analysis and pathway enrichment},
  author={Cho, Hyun Jae and Shu, Mia and Bekiranov, Stefan and Zang, Chongzhi and Zhang, Aidong},
  journal={Bioinformatics},
  volume={39},
  number={4},
  pages={btad113},
  year={2023},
  publisher={Oxford University Press}
}

@article{cottin2022idnetwork,
  title={IDNetwork: A deep illness-death network based on multi-state event history process for disease prognostication},
  author={Cottin, Aziliz and Pecuchet, Nicolas and Zulian, Marine and Guilloux, Agathe and Katsahian, Sandrine},
  journal={Statistics in Medicine},
  volume={41},
  number={9},
  pages={1573--1598},
  year={2022},
  publisher={Wiley Online Library}
}

@article{cox1972regression,
  title={Regression models and life-tables},
  author={Cox, David R},
  journal={Journal of the Royal Statistical Society: Series B (Methodological)},
  volume={34},
  number={2},
  pages={187--202},
  year={1972},
  publisher={Wiley Online Library}
}

@article{deepa2022systematic,
  title={A systematic review on machine learning and deep learning techniques in cancer survival prediction},
  author={Deepa, P and Gunavathi, C},
  journal={Progress in Biophysics and Molecular Biology},
  year={2022},
  publisher={Elsevier}
}

@article{fan2022survival,
  title={Survival Analysis with High-Dimensional Omics Data Using a Threshold Gradient Descent Regularization-Based Neural Network Approach},
  author={Fan, Yu and Zhang, Sanguo and Ma, Shuangge},
  journal={Genes},
  volume={13},
  number={9},
  pages={1674},
  year={2022},
  publisher={MDPI}
}

@article{faraggi1995neural,
  title={A neural network model for survival data},
  author={Faraggi, David and Simon, Richard},
  journal={Statistics in medicine},
  volume={14},
  number={1},
  pages={73--82},
  year={1995},
  publisher={Wiley Online Library}
}

@inproceedings{fornili2013piecewise,
  title={Piecewise exponential artificial neural networks (PEANN) for modeling hazard function with right censored data},
  author={Fornili, Marco and Ambrogi, Federico and Boracchi, Patrizia and Biganzoli, Elia},
  booktitle={International Meeting on Computational Intelligence Methods for Bioinformatics and Biostatistics},
  pages={125--136},
  year={2013},
  publisher={Springer}
}

@article{friedman1982piecewise,
  title={Piecewise exponential models for survival data with covariates},
  author={Friedman, Michael},
  journal={The Annals of Statistics},
  volume={10},
  number={1},
  pages={101--113},
  year={1982},
  publisher={Institute of Mathematical Statistics}
}

@inproceedings{fuhlert2022deep,
  title={Deep Learning-Based Discrete Calibrated Survival Prediction},
  author={Fuhlert, Patrick and Ernst, Anne and Dietrich, Esther and Westhaeusser, Fabian and Kloiber, Karin and Bonn, Stefan},
  booktitle={2022 IEEE International Conference on Digital Health (ICDH)},
  pages={169--174},
  year={2022},
  organization={IEEE}
}

@article{gensheimer2019scalable,
  title={A scalable discrete-time survival model for neural networks},
  author={Gensheimer, Michael F and Narasimhan, Balasubramanian},
  journal={PeerJ},
  volume={7},
  pages={e6257},
  year={2019},
  publisher={PeerJ Inc.}
}

@inproceedings{goodfellow2014generative,
  title={Generative Adversarial Nets},
  author={Goodfellow, Ian J and Pouget-Abadie, Jean and Mirza, Mehdi and Xu, Bing and Warde-Farley, David and Ozair, Sherjil and Courville, Aaron C and Bengio, Yoshua},
  booktitle={NIPS},
  year={2014}
}

@inproceedings{gupta2019cresa,
  title={Cresa: a deep learning approach to competing risks, recurrent event survival analysis},
  author={Gupta, Garima and Sunder, Vishal and Prasad, Ranjitha and Shroff, Gautam},
  booktitle={Advances in Knowledge Discovery and Data Mining: 23rd Pacific-Asia Conference, PAKDD 2019, Macau, China, April 14-17, 2019, Proceedings, Part II 23},
  pages={108--122},
  year={2019},
  organization={Springer}
}

@inproceedings{hao2019page,
  title={PAGE-Net: interpretable and integrative deep learning for survival analysis using histopathological images and genomic data},
  author={Hao, Jie and Kosaraju, Sai Chandra and Tsaku, Nelson Zange and Song, Dae Hyun and Kang, Mingon},
  booktitle={Pacific Symposium on Biocomputing 2020},
  pages={355--366},
  year={2019},
  organization={World Scientific}
}

@inproceedings{he2016deep,
  title={Deep residual learning for image recognition},
  author={He, Kaiming and Zhang, Xiangyu and Ren, Shaoqing and Sun, Jian},
  booktitle={Proceedings of the IEEE conference on computer vision and pattern recognition},
  pages={770--778},
  year={2016}
}

@article{herrmann2021large,
  title={Large-scale benchmark study of survival prediction methods using multi-omics data},
  author={Herrmann, Moritz and Probst, Philipp and Hornung, Roman and Jurinovic, Vindi and Boulesteix, Anne-Laure},
  journal={Briefings in bioinformatics},
  volume={22},
  number={3},
  pages={bbaa167},
  year={2021},
  publisher={Oxford University Press}
}

@article{huang2019salmon,
  title={SALMON: survival analysis learning with multi-omics neural networks on breast cancer},
  author={Huang, Zhi and Zhan, Xiaohui and Xiang, Shunian and Johnson, Travis S and Helm, Bryan and Yu, Christina Y and Zhang, Jie and Salama, Paul and Rizkalla, Maher and Han, Zhi and others},
  journal={Frontiers in genetics},
  volume={10},
  pages={166},
  year={2019},
  publisher={Frontiers Media SA}
}

@inproceedings{irvin2019chexpert,
  title={Chexpert: A large chest radiograph dataset with uncertainty labels and expert comparison},
  author={Irvin, Jeremy and Rajpurkar, Pranav and Ko, Michael and Yu, Yifan and Ciurea-Ilcus, Silviana and Chute, Chris and Marklund, Henrik and Haghgoo, Behzad and Ball, Robyn and Shpanskaya, Katie and others},
  booktitle={Proceedings of the AAAI conference on artificial intelligence},
  volume={33},
  number={01},
  pages={590--597},
  year={2019}
}

@article{ishwaran2008random,
  title={Random survival forests},
  author={Ishwaran, Hemant and Kogalur, Udaya B and Blackstone, Eugene H and Lauer, Michael S and others},
  journal={The annals of applied statistics},
  volume={2},
  number={3},
  pages={841--860},
  year={2008},
  publisher={Institute of Mathematical Statistics}
}

@book{ivakhnenko1967cybernetics,
  title={Cybernetics and forecasting techniques},
  author = {Ivakhnenko, Alekse{\u\i} Grigor'evich and Lapa, Valentin Grigorevich and Lapa, Valentin Grigor'evich},
  volume={8},
  year={1967},
  publisher={American Elsevier Publishing Company}
}

@inproceedings{jetley2018learn,
  title={Learn to pay attention},
  author={Jetley, S and Lord, NA and Lee, N and Torr, PH},
  booktitle={International Conference on Learning Representations},
  year={2018},
  organization={International Conference on Learning Representations}
}

@article{jing2019deep,
  title={A deep survival analysis method based on ranking},
  author={Jing, Bingzhong and Zhang, Tao and Wang, Zixian and Jin, Ying and Liu, Kuiyuan and Qiu, Wenze and Ke, Liangru and Sun, Ying and He, Caisheng and Hou, Dan and others},
  journal={Artificial intelligence in medicine},
  volume={98},
  pages={1--9},
  year={2019},
  publisher={Elsevier}
}

@article{kalakoti2021survcnn,
  title={SurvCNN: a discrete time-to-event cancer survival estimation framework using image representations of omics data},
  author={Kalakoti, Yogesh and Yadav, Shashank and Sundar, Durai},
  journal={Cancers},
  volume={13},
  number={13},
  pages={3106},
  year={2021},
  publisher={MDPI}
}

@book{kalbfleisch2011statistical,
  title={The statistical analysis of failure time data},
  author={Kalbfleisch, John D and Prentice, Ross L},
  year={2011},
  publisher={John Wiley \& Sons}
}

@inproceedings{kamran2021estimating,
  title={Estimating calibrated individualized survival curves with deep learning},
  author={Kamran, Fahad and Wiens, Jenna},
  booktitle={Proceedings of the AAAI Conference on Artificial Intelligence},
  volume={35},
  number={1},
  pages={240--248},
  year={2021}
}

@article{kaplan1958nonparametric,
  title={Nonparametric estimation from incomplete observations},
  author={Kaplan, Edward L and Meier, Paul},
  journal={Journal of the American statistical association},
  volume={53},
  number={282},
  pages={457--481},
  year={1958},
  publisher={Taylor \& Francis}
}

@article{katzman2018deepsurv,
  title={DeepSurv: personalized treatment recommender system using a Cox proportional hazards deep neural network},
  author={Katzman, Jared L and Shaham, Uri and Cloninger, Alexander and Bates, Jonathan and Jiang, Tingting and Kluger, Yuval},
  journal={BMC medical research methodology},
  volume={18},
  number={1},
  pages={24},
  year={2018},
  publisher={Springer}
}

@inproceedings{2015-kingma,
  author = {Kingma, Diederik P. and Ba, Jimmy},
  booktitle = {ICLR (Poster)},
  editor = {Bengio, Yoshua and LeCun, Yann},
  ee = {http://arxiv.org/abs/1412.6980},
  title = {Adam: A Method for Stochastic Optimization.},
  url = {http://dblp.uni-trier.de/db/conf/iclr/iclr2015.html#KingmaB14},
  year = 2015
}

@inproceedings{kopper2022deeppamm,
  title={DeepPAMM: Deep Piecewise Exponential Additive Mixed Models for Complex Hazard Structures in Survival Analysis},
  author={Kopper, Philipp and Wiegrebe, Simon and Bischl, Bernd and Bender, Andreas and R{\"u}gamer, David},
  booktitle={Advances in Knowledge Discovery and Data Mining: 26th Pacific-Asia Conference, PAKDD 2022, Chengdu, China, May 16--19, 2022, Proceedings, Part II},
  pages={249--261},
  year={2022},
  organization={Springer}
}

@article{krzyzinski2022survshap,
  title={SurvSHAP (t): Time-dependent explanations of machine learning survival models},
  author={Krzyzi{\'n}ski, Mateusz and Spytek, Miko{\l}aj and Baniecki, Hubert and Biecek, Przemys{\l}aw},
  journal={Knowledge-Based Systems},
  pages={110234},
  year={2022},
  publisher={Elsevier}
}

@article{kvamme2019time,
  title={Time-to-event prediction with neural networks and Cox regression},
  author={Kvamme, H{\aa}vard and Borgan, {\O}rnulf and Scheel, Ida},
  journal={Journal of machine learning research},
  volume={20},
  number={129},
  pages={1--30},
  year={2019}
}

@article{kvamme2021continuous,
  title={Continuous and discrete-time survival prediction with neural networks},
  author={Kvamme, H{\aa}vard and Borgan, {\O}rnulf},
  journal={Lifetime data analysis},
  volume={27},
  pages={710--736},
  year={2021},
  publisher={Springer}
}

@article{lecun1989backpropagation,
  title={Backpropagation applied to handwritten zip code recognition},
  author={LeCun, Yann and Boser, Bernhard and Denker, John S and Henderson, Donnie and Howard, Richard E and Hubbard, Wayne and Jackel, Lawrence D},
  journal={Neural computation},
  volume={1},
  number={4},
  pages={541--551},
  year={1989},
  publisher={MIT Press}
}

@article{lee2019review,
  title={Review of statistical methods for survival analysis using genomic data},
  author={Lee, Seungyeoun and Lim, Heeju},
  journal={Genomics \& informatics},
  volume={17},
  number={4},
  year={2019},
  publisher={Korea Genome Organization}
}

@article{lee1975fuzzy,
  title={Fuzzy neural networks},
  author={Lee, Samuel C and Lee, Edward T},
  journal={Mathematical Biosciences},
  volume={23},
  number={1-2},
  pages={151--177},
  year={1975},
  publisher={Elsevier}
}

@article{lee2019dynamic,
  title={Dynamic-Deephit: A deep learning approach for dynamic survival analysis with competing risks based on longitudinal data},
  author={Lee, Changhee and Yoon, Jinsung and Van Der Schaar, Mihaela},
  journal={IEEE Transactions on Biomedical Engineering},
  volume={67},
  number={1},
  pages={122--133},
  year={2019},
  publisher={IEEE}
}

@article{liestbl1994survival,
  title={Survival analysis and neural nets},
  author={Liestbl, Knut and Andersen, Per Kragh and Andersen, Ulrich},
  journal={Statistics in medicine},
  volume={13},
  number={12},
  pages={1189--1200},
  year={1994},
  publisher={Wiley Online Library}
}

@article{liu2022deep,
  title={Deep learning for survival analysis in breast cancer with whole slide image data},
  author={Liu, Huidong and Kurc, Tahsin},
  journal={Bioinformatics},
  volume={38},
  number={14},
  pages={3629--3637},
  year={2022},
  publisher={Oxford University Press}
}

@inproceedings{lee2018deephit,
  title={DeepHit: A Deep Learning Approach to Survival Analysis With Competing Risks.},
  author={Lee, Changhee and Zame, William R and Yoon, Jinsung and van der Schaar, Mihaela},
  booktitle={AAAI},
  pages={2314--2321},
  year={2018}
}

@article{lin2022deep,
  title={Deep learning for the dynamic prediction of multivariate longitudinal and survival data},
  author={Lin, Jeffrey and Luo, Sheng},
  journal={Statistics in medicine},
  volume={41},
  number={15},
  pages={2894--2907},
  year={2022},
  publisher={Wiley Online Library}
}

@article{lundberg2017unified,
  title={A unified approach to interpreting model predictions},
  author={Lundberg, Scott M and Lee, Su-In},
  journal={Advances in neural information processing systems},
  volume={30},
  year={2017}
}

@article{montavon2019layer,
  title={Layer-wise relevance propagation: an overview},
  author={Montavon, Gr{\'e}goire and Binder, Alexander and Lapuschkin, Sebastian and Samek, Wojciech and M{\"u}ller, Klaus-Robert},
  journal={Explainable AI: interpreting, explaining and visualizing deep learning},
  pages={193--209},
  year={2019},
  publisher={Springer}
}

@article{meister2008statistical,
  title={Statistical methods for estimating the probability of spontaneous abortion in observational studies—analyzing pregnancies exposed to coumarin derivatives},
  author={Meister, Reinhard and Schaefer, Christof},
  journal={Reproductive Toxicology},
  volume={26},
  number={1},
  pages={31--35},
  year={2008},
  publisher={Elsevier}
}

@inproceedings{nagpal2021deep_dcm,
  title={Deep Cox mixtures for survival regression},
  author={Nagpal, Chirag and Yadlowsky, Steve and Rostamzadeh, Negar and Heller, Katherine},
  booktitle={Machine Learning for Healthcare Conference},
  pages={674--708},
  year={2021},
  organization={PMLR}
}

@article{nagpal2021_dsm,
  title={Deep survival machines: Fully parametric survival regression and representation learning for censored data with competing risks},
  author={Nagpal, Chirag and Li, Xinyu and Dubrawski, Artur},
  journal={IEEE Journal of Biomedical and Health Informatics},
  volume={25},
  number={8},
  pages={3163--3175},
  year={2021},
  key={Nagpal2021a},
  publisher={IEEE}
}

@inproceedings{nagpal2021_rdsm,
  title={Deep parametric time-to-event regression with time-varying covariates},
  author={Nagpal, Chirag and Jeanselme, Vincent and Dubrawski, Artur},
  booktitle={Survival Prediction-Algorithms, Challenges and Applications},
  pages={184--193},
  year={2021},
  key={Nagpal2021b},
  organization={PMLR}
}

@article{auton-survival,
  title={auton-survival: an Open-Source Package for Regression, Counterfactual Estimation, Evaluation and Phenotyping with Censored Time-to-Event Data},
  author={Nagpal, Chirag and Potosnak, Willa and Dubrawski, Artur},
  year={2022},
  journal={arXiv preprint arXiv:2204.07276}  ,
  publisher={arXiv}
}

@inproceedings{nagpal2022auton,
  title={auton-survival: An open-source package for regression, counterfactual estimation, evaluation and phenotyping with censored time-to-event data},
  author={Nagpal, Chirag and Potosnak, Willa and Dubrawski, Artur},
  booktitle={Machine Learning for Healthcare Conference},
  pages={585--608},
  year={2022},
  organization={PMLR}
}

@article{noordzij2013we,
  title={When do we need competing risks methods for survival analysis in nephrology?},
  author={Noordzij, Marlies and Leffondr{\'e}, Karen and van Stralen, Karlijn J and Zoccali, Carmine and Dekker, Friedo W and Jager, Kitty J},
  journal={Nephrology Dialysis Transplantation},
  volume={28},
  number={11},
  pages={2670--2677},
  year={2013},
  publisher={Oxford University Press}
}

@article{ssdr,
author = {David Rügamer and Chris Kolb and Nadja Klein},
title = {Semi-Structured Distributional Regression},
journal = {The American Statistician},
volume = {0},
number = {0},
pages = {1-12},
year  = {2023},
publisher = {Taylor & Francis},
doi = {10.1080/00031305.2022.2164054},
URL = {https://doi.org/10.1080/00031305.2022.2164054},
eprint = {https://doi.org/10.1080/00031305.2022.2164054}
}

@article{sksurv,
  author  = {Sebastian P{\"o}lsterl},
  title   = {scikit-survival: A Library for Time-to-Event Analysis Built on Top of scikit-learn},
  journal = {Journal of Machine Learning Research},
  year    = {2020},
  volume  = {21},
  number  = {212},
  pages   = {1-6},
  url     = {http://jmlr.org/papers/v21/20-729.html}
}

@inproceedings{qi2017pointnet,
  title={Pointnet: Deep learning on point sets for 3d classification and segmentation},
  author={Qi, Charles R and Su, Hao and Mo, Kaichun and Guibas, Leonidas J},
  booktitle={Proceedings of the IEEE conference on computer vision and pattern recognition},
  pages={652--660},
  year={2017}
}

@article{qiu2020meta,
  title={A meta-learning approach for genomic survival analysis},
  author={Qiu, Yeping Lina and Zheng, Hong and Devos, Arnout and Selby, Heather and Gevaert, Olivier},
  journal={Nature communications},
  volume={11},
  number={1},
  pages={6350},
  year={2020},
  publisher={Nature Publishing Group UK London}
}

@article{qin2022survival,
  title={Survival prediction model for right-censored data based on improved composite quantile regression neural network.},
  author={Qin, Xiwen and Yin, Dongmei and Dong, Xiaogang and Chen, Dongxue and Zhang, Shuang},
  journal={Mathematical Biosciences and Engineering: MBE},
  volume={19},
  number={8},
  pages={7521--7542},
  year={2022}
}

@article{ramjith2021flexible,
  title={Flexible modelling of risk factors on the incidence of pneumonia in young children in South Africa using piece-wise exponential additive mixed modelling},
  author={Ramjith, Jordache and Roes, Kit CB and Zar, Heather J and Jonker, Marianne A},
  journal={BMC Medical Research Methodology},
  volume={21},
  number={1},
  pages={1--13},
  year={2021},
  publisher={Springer}
}

@Article{mlr3proba,
  title = {mlr3proba: An R Package for Machine Learning in Survival Analysis},
  author = {Raphael Sonabend and Franz J Király and Andreas Bender and Bernd Bischl and Michel Lang},
  journal = {Bioinformatics},
  month = {02},
  year = {2021},
  doi = {10.1093/bioinformatics/btab039},
  issn = {1367-4803},
}

@inproceedings{ren2019deep,
  title={Deep recurrent survival analysis},
  author={Ren, Kan and Qin, Jiarui and Zheng, Lei and Yang, Zhengyu and Zhang, Weinan and Qiu, Lin and Yu, Yong},
  booktitle={Proceedings of the AAAI Conference on Artificial Intelligence},
  volume={33},
  number={01},
  pages={4798--4805},
  year={2019}
}

@inproceedings{rezende2015variational,
  title={Variational inference with normalizing flows},
  author={Rezende, Danilo and Mohamed, Shakir},
  booktitle={International conference on machine learning},
  pages={1530--1538},
  year={2015},
  organization={PMLR}
}

@inproceedings{ribeiro2016should,
  title={" Why should i trust you?" Explaining the predictions of any classifier},
  author={Ribeiro, Marco Tulio and Singh, Sameer and Guestrin, Carlos},
  booktitle={Proceedings of the 22nd ACM SIGKDD international conference on knowledge discovery and data mining},
  pages={1135--1144},
  year={2016}
}

@article{rosenblatt1967recent,
  title={Recent work on theoretical models of biological memory},
  author={Rosenblatt, Frank},
  journal={Computer and information sciences II},
  pages={33--56},
  year={1967}
}

@article{ruder2017overview,
  title={An overview of multi-task learning in deep neural networks},
  author={Ruder, Sebastian},
  journal={arXiv preprint arXiv:1706.05098},
  year={2017}
}

@article{rumelhart1986learning,
  title={Learning representations by back-propagating errors},
  author={Rumelhart, David E and Hinton, Geoffrey E and Williams, Ronald J},
  journal={nature},
  volume={323},
  number={6088},
  pages={533--536},
  year={1986},
  publisher={Nature Publishing Group}
}

@article{sabour2017dynamic,
  title={Dynamic routing between capsules},
  author={Sabour, Sara and Frosst, Nicholas and Hinton, Geoffrey E},
  journal={Advances in neural information processing systems},
  volume={30},
  year={2017}
}

@inproceedings{sansaengtham2020survival,
  title={Survival Analysis For Computing Systems Using A Deep Ensemble Network},
  author={Sansaengtham, Baramee and Barroso, Vasco Chibante and Phunchongharn, Phond},
  booktitle={2020 IEEE 6th International Conference on Control Science and Systems Engineering (ICCSSE)},
  pages={57--62},
  year={2020},
  organization={IEEE}
}

@article{schwarzer2000misuses,
  title={On the misuses of artificial neural networks for prognostic and diagnostic classification in oncology},
  author={Schwarzer, Guido and Vach, Werner and Schumacher, Martin},
  journal={Statistics in medicine},
  volume={19},
  number={4},
  pages={541--561},
  year={2000},
  publisher={Wiley Online Library}
}

@article{shin2019cascaded,
  title={Cascaded Wx: A novel prognosis-related feature selection framework in human lung adenocarcinoma transcriptomes},
  author={Shin, Bonggun and Park, Sungsoo and Hong, Ji Hyung and An, Ho Jung and Chun, Sang Hoon and Kang, Kilsoo and Ahn, Young-Ho and Ko, Yoon Ho and Kang, Keunsoo},
  journal={Frontiers in Genetics},
  volume={10},
  pages={662},
  year={2019},
  publisher={Frontiers Media SA}
}

@inproceedings{shrikumar2017learning,
  title={Learning important features through propagating activation differences},
  author={Shrikumar, Avanti and Greenside, Peyton and Kundaje, Anshul},
  booktitle={International conference on machine learning},
  pages={3145--3153},
  year={2017},
  organization={PMLR}
}

@article{steele2018machine,
  title={Machine learning models in electronic health records can outperform conventional survival models for predicting patient mortality in coronary artery disease},
  author={Steele, Andrew J and Denaxas, Spiros C and Shah, Anoop D and Hemingway, Harry and Luscombe, Nicholas M},
  journal={PloS one},
  volume={13},
  number={8},
  pages={e0202344},
  year={2018},
  publisher={Public Library of Science San Francisco, CA USA}
}

@inproceedings{sohl2015deep,
  title={Deep Unsupervised Learning using Nonequilibrium Thermodynamics},
  author={Sohl-Dickstein, Jascha and Weiss, Eric A and Maheswaranathan, Niru and Ganguli, Surya},
  booktitle={ICML},
  year={2015}
}

@article{sun2020genome,
  title={Genome-wide association study-based deep learning for survival prediction},
  author={Sun, Tao and Wei, Yue and Chen, Wei and Ding, Ying},
  journal={Statistics in medicine},
  volume={39},
  number={30},
  pages={4605--4620},
  year={2020},
  publisher={Wiley Online Library}
}

@article{utkin2022survnam,
  title={SurvNAM: The machine learning survival model explanation},
  author={Utkin, Lev V and Satyukov, Egor D and Konstantinov, Andrei V},
  journal={Neural Networks},
  volume={147},
  pages={81--102},
  year={2022},
  publisher={Elsevier}
}

@article{thorsen2022discrete,
  title={Discrete-time survival analysis in the critically ill: a deep learning approach using heterogeneous data},
  author={Thorsen-Meyer, Hans-Christian and Placido, Davide and Kaas-Hansen, Benjamin Skov and Nielsen, Anna P and Lange, Theis and Nielsen, Annelaura B and Toft, Palle and Schierbeck, Jens and Str{\o}m, Thomas and Chmura, Piotr J and others},
  journal={NPJ digital medicine},
  volume={5},
  number={1},
  pages={142},
  year={2022},
  publisher={Nature Publishing Group UK London}
}

@article{tong2022deep,
  title={Deep survival algorithm based on nuclear norm},
  author={Tong, Jianyang and Zhao, Xuejing},
  journal={Journal of Statistical Computation and Simulation},
  volume={92},
  number={9},
  pages={1964--1976},
  year={2022},
  publisher={Taylor \& Francis}
}

@book{tutz2016modeling,
  title={Modeling discrete time-to-event data},
  author={Tutz, Gerhard and Schmid, Matthias and others},
  year={2016},
  publisher={Springer}
}

@article{vale2021multisurv,
  title={Long-term cancer survival prediction using multimodal deep learning},
  author={Vale-Silva, Lu{\'\i}s A and Rohr, Karl},
  journal={Scientific Reports},
  volume={11},
  number={1},
  pages={13505},
  year={2021},
  publisher={Nature Publishing Group UK London}
}

@inproceedings{vaswani2017attention_2,
author = {Vaswani, Ashish and Shazeer, Noam and Parmar, Niki and Uszkoreit, Jakob and Jones, Llion and Gomez, Aidan N. and Kaiser, \L{}ukasz and Polosukhin, Illia},
title = {Attention is all you need},
year = {2017},
isbn = {9781510860964},
publisher = {Curran Associates Inc.},
address = {Red Hook, NY, USA},
booktitle = {Proceedings of the 31st International Conference on Neural Information Processing Systems},
pages = {6000–6010},
numpages = {11},
location = {Long Beach, California, USA},
series = {NIPS'17}
}

@article{vellido2020importance,
  title={The importance of interpretability and visualization in machine learning for applications in medicine and health care},
  author={Vellido, Alfredo},
  journal={Neural computing and applications},
  volume={32},
  number={24},
  pages={18069--18083},
  year={2020},
  publisher={Springer}
}

@article{wang2019machine,
  title={Machine learning for survival analysis: A survey},
  author={Wang, Ping and Li, Yan and Reddy, Chandan K},
  journal={ACM Computing Surveys (CSUR)},
  volume={51},
  number={6},
  pages={1--36},
  year={2019},
  publisher={ACM New York, NY, USA}
}

@article{wang2019extreme,
  title={Extreme learning machine Cox model for high-dimensional survival analysis},
  author={Wang, Hong and Li, Gang},
  journal={Statistics in medicine},
  volume={38},
  number={12},
  pages={2139--2156},
  year={2019},
  publisher={Wiley Online Library}
}

@inproceedings{wang2022survtrace,
  title={Survtrace: Transformers for survival analysis with competing events},
  author={Wang, Zifeng and Sun, Jimeng},
  booktitle={Proceedings of the 13th ACM International Conference on Bioinformatics, Computational Biology and Health Informatics},
  pages={1--9},
  year={2022}
}

@article{wijethilake2021glioma,
  title={Glioma survival analysis empowered with data engineering—a survey},
  author={Wijethilake, Navodini and Meedeniya, Dulani and Chitraranjan, Charith and Perera, Indika and Islam, Mobarakol and Ren, Hongliang},
  journal={Ieee Access},
  volume={9},
  pages={43168--43191},
  year={2021},
  publisher={IEEE}
}

@article{wolf2022daft,
  title={DAFT: A universal module to interweave tabular data and 3D images in CNNs},
  author={Wolf, Tom Nuno and P{\"o}lsterl, Sebastian and Wachinger, Christian and Alzheimer’s Disease Neuroimaging Initiative and others},
  journal={NeuroImage},
  volume={260},
  pages={119505},
  year={2022},
  publisher={Elsevier}
}

@article{wolpert1992stacked,
  title={Stacked generalization},
  author={Wolpert, David H},
  journal={Neural networks},
  volume={5},
  number={2},
  pages={241--259},
  year={1992},
  publisher={Elsevier}
}

@article{wu2019selective,
  title={A selective review of multi-level omics data integration using variable selection},
  author={Wu, Cen and Zhou, Fei and Ren, Jie and Li, Xiaoxi and Jiang, Yu and Ma, Shuangge},
  journal={High-throughput},
  volume={8},
  number={1},
  pages={4},
  year={2019},
  publisher={MDPI}
}

@article{xie2019group,
  title={Group lasso regularized deep learning for cancer prognosis from multi-omics and clinical features},
  author={Xie, Gangcai and Dong, Chengliang and Kong, Yinfei and Zhong, Jiang F and Li, Mingyao and Wang, Kai},
  journal={Genes},
  volume={10},
  number={3},
  pages={240},
  year={2019},
  publisher={MDPI}
}

@article{yin2022convolutional,
  title={A convolutional neural network model for survival prediction based on prognosis-related cascaded Wx feature selection},
  author={Yin, Qingyan and Chen, Wangwang and Zhang, Chunxia and Wei, Zhi},
  journal={Laboratory Investigation},
  volume={102},
  number={10},
  pages={1064--1074},
  year={2022},
  publisher={Nature Publishing Group US New York}
}

@article{yousefi2017predicting,
  title={Predicting clinical outcomes from large scale cancer genomic profiles with deep survival models},
  author={Yousefi, Safoora and Amrollahi, Fatemeh and Amgad, Mohamed and Dong, Chengliang and Lewis, Joshua E and Song, Congzheng and Gutman, David A and Halani, Sameer H and Velazquez Vega, Jose Enrique and Brat, Daniel J and others},
  journal={Scientific reports},
  volume={7},
  number={1},
  pages={11707},
  year={2017},
  publisher={Nature Publishing Group UK London}
}

@inproceedings{yu2011learning_2,
author = {Yu, Chun-Nam and Greiner, Russell and Lin, Hsiu-Chin and Baracos, Vickie},
title = {Learning patient-specific cancer survival distributions as a sequence of dependent regressors},
year = {2011},
isbn = {9781618395993},
publisher = {Curran Associates Inc.},
address = {Red Hook, NY, USA},
booktitle = {Proceedings of the 24th International Conference on Neural Information Processing Systems},
pages = {1845–1853},
numpages = {9},
location = {Granada, Spain},
series = {NIPS'11}
}

@article{zhang2014normalized,
  title={Normalized imqcm: An algorithm for detecting weak quasi-cliques in weighted graph with applications in gene co-expression module discovery in cancers},
  author={Zhang, Jie and Huang, Kun},
  journal={Cancer informatics},
  volume={13},
  pages={CIN--S14021},
  year={2014},
  publisher={SAGE Publications Sage UK: London, England}
}

@article{zhang2020cnn,
  title={CNN-based survival model for pancreatic ductal adenocarcinoma in medical imaging},
  author={Zhang, Yucheng and Lobo-Mueller, Edrise M and Karanicolas, Paul and Gallinger, Steven and Haider, Masoom A and Khalvati, Farzad},
  journal={BMC medical imaging},
  volume={20},
  pages={1--8},
  year={2020},
  publisher={Springer}
}

@article{zhao2021deepomix,
  title={DeepOmix: A scalable and interpretable multi-omics deep learning framework and application in cancer survival analysis},
  author={Zhao, Lianhe and Dong, Qiongye and Luo, Chunlong and Wu, Yang and Bu, Dechao and Qi, Xiaoning and Luo, Yufan and Zhao, Yi},
  journal={Computational and structural biotechnology journal},
  volume={19},
  pages={2719--2725},
  year={2021},
  publisher={Elsevier}
}

@article{zhang2022survbenchmark,
  title={SurvBenchmark: comprehensive benchmarking study of survival analysis methods using both omics data and clinical data},
  author={Zhang, Yunwei and Wong, Germaine and Mann, Graham and Muller, Samuel and Yang, Jean YH},
  journal={GigaScience},
  volume={11},
  year={2022},
  publisher={Oxford Academic}
}

@inproceedings{kopper2021semi,
  title={Semi-structured deep piecewise exponential models},
  author={Kopper, Philipp and P{\"o}lsterl, Sebastian and Wachinger, Christian and Bischl, Bernd and Bender, Andreas and R{\"u}gamer, David},
  booktitle={Survival Prediction-Algorithms, Challenges and Applications},
  pages={40--53},
  year={2021},
  organization={PMLR}
}

@inproceedings{bender_general_2021,
	address = {Cham},
	series = {Lecture {Notes} in {Computer} {Science}},
	title = {A {General} {Machine} {Learning} {Framework} for {Survival} {Analysis}},
	isbn = {978-3-030-67664-3},
	doi = {10.1007/978-3-030-67664-3_10},
	language = {en},
	booktitle = {Machine {Learning} and {Knowledge} {Discovery} in {Databases}},
	publisher = {Springer International Publishing},
	author = {Bender, Andreas and Rügamer, David and Scheipl, Fabian and Bischl, Bernd},
	editor = {Hutter, Frank and Kersting, Kristian and Lijffijt, Jefrey and Valera, Isabel},
	month = feb,
	year = {2021},
	pages = {158--173},
}

@article{groha_general_2021,
	title = {A {General} {Framework} for {Survival} {Analysis} and {Multi}-{State} {Modelling}},
	url = {http://arxiv.org/abs/2006.04893},
	urldate = {2021-03-20},
	journal = {arXiv:2006.04893  [cs, stat]},
	author = {Groha, Stefan and Schmon, Sebastian M. and Gusev, Alexander},
	month = feb,
	year = {2021},
	note = {arXiv: 2006.04893},
}

@article{zhao_dnnsurv_2019,
	title = {{DNNSurv}: {Deep} {Neural} {Networks} for {Survival} {Analysis} {Using} {Pseudo} {Values}},
	shorttitle = {{DNNSurv}},
	url = {http://arxiv.org/abs/1908.02337},
	urldate = {2021-03-20},
	journal = {arXiv:1908.02337 [cs, stat]},
	author = {Zhao, Lili and Feng, Dai},
	month = aug,
	year = {2019},
	note = {arXiv: 1908.02337
version: 1},
}

@article{fotso_deep_2018,
	title = {Deep {Neural} {Networks} for {Survival} {Analysis} {Based} on a {Multi}-{Task} {Framework}},
	url = {http://arxiv.org/abs/1801.05512},
	urldate = {2021-03-21},
	journal = {arXiv:1801.05512   [cs, stat]},
	author = {Fotso, Stephane},
	month = jan,
	year = {2018},
	note = {arXiv: 1801.05512},
}

@article{kvamme_continuous_2019,
	title = {Continuous and {Discrete}-{Time} {Survival} {Prediction} with {Neural} {Networks}},
	url = {http://arxiv.org/abs/1910.06724},
	urldate = {2019-11-19},
	journal = {arXiv:1910.06724 [cs, stat]},
	author = {Kvamme, Håvard and Borgan, {\O}rnulf},
	month = oct,
	year = {2019},
	note = {arXiv: 1910.06724},
}

@inproceedings{polsterl_wide_2020,
	address = {Cham},
	series = {Communications in {Computer} and {Information} {Science}},
	title = {A {Wide} and {Deep} {Neural} {Network} for {Survival} {Analysis} from {Anatomical} {Shape} and {Tabular} {Clinical} {Data}},
	isbn = {978-3-030-43823-4},
	doi = {10.1007/978-3-030-43823-4_37},
	language = {en},
	booktitle = {Machine {Learning} and {Knowledge} {Discovery} in {Databases}},
	publisher = {Springer International Publishing},
	author = {Pölsterl, Sebastian and Sarasua, Ignacio and Gutiérrez-Becker, Benjamín and Wachinger, Christian},
	editor = {Cellier, Peggy and Driessens, Kurt},
	year = {2020},
	pages = {453--464},
}

@inproceedings{giunchiglia_rnn-surv_2018,
	address = {Cham},
	series = {Lecture {Notes} in {Computer} {Science}},
	title = {{RNN}-{SURV}: {A} {Deep} {Recurrent} {Model} for {Survival} {Analysis}},
	isbn = {978-3-030-01424-7},
	shorttitle = {{RNN}-{SURV}},
	doi = {10.1007/978-3-030-01424-7_3},
	language = {en},
	booktitle = {Artificial {Neural} {Networks} and {Machine} {Learning} – {ICANN} 2018},
	publisher = {Springer International Publishing},
	author = {Giunchiglia, Eleonora and Nemchenko, Anton and van der Schaar, Mihaela},
	editor = {Kůrková, Věra and Manolopoulos, Yannis and Hammer, Barbara and Iliadis, Lazaros and Maglogiannis, Ilias},
	year = {2018},
	pages = {23--32},
}

@book{klein_survival_1997,
	address = {New York},
	title = {Survival analysis: {Techniques} for censored and truncated data},
	isbn = {978-0-387-94829-4},
	publisher = {Springer},
	author = {Klein, John P. and Moeschberger, Melvin L.},
	year = {1997},
}

@inproceedings{hu_transformer-based_2021,
	title = {Transformer-{Based} {Deep} {Survival} {Analysis}},
	url = {https://proceedings.mlr.press/v146/hu21a.html},
	language = {en},
	urldate = {2021-10-31},
	booktitle = {Proceedings of {AAAI} {Spring} {Symposium} on {Survival} {Prediction} - {Algorithms}, {Challenges}, and {Applications} 2021},
	publisher = {PMLR},
	author = {Hu, Shi and Fridgeirsson, Egill and Wingen, Guido van and Welling, Max},
	month = may,
	year = {2021},
	note = {ISSN: 2640-3498},
	pages = {132--148},
}

@inproceedings{bennis_dpwte_2021,
	address = {Cham},
	series = {Lecture {Notes} in {Computer} {Science}},
	title = {{DPWTE}: {A} {Deep} {Learning} {Approach} to {Survival} {Analysis} {Using} a {Parsimonious} {Mixture} of {Weibull} {Distributions}},
	isbn = {978-3-030-86340-1},
	shorttitle = {{DPWTE}},
	doi = {10.1007/978-3-030-86340-1_15},
	language = {en},
	booktitle = {Artificial {Neural} {Networks} and {Machine} {Learning} – {ICANN} 2021},
	publisher = {Springer International Publishing},
	author = {Bennis, Achraf and Mouysset, Sandrine and Serrurier, Mathieu},
	editor = {Farkaš, Igor and Masulli, Paolo and Otte, Sebastian and Wermter, Stefan},
	year = {2021},
	pages = {185--196},
}

@article{nezhad_deep_2019,
	title = {A {Deep} {Active} {Survival} {Analysis} approach for precision treatment recommendations: {Application} of prostate cancer},
	volume = {115},
	issn = {0957-4174},
	shorttitle = {A {Deep} {Active} {Survival} {Analysis} approach for precision treatment recommendations},
	url = {https://www.sciencedirect.com/science/article/pii/S0957417418304949},
	doi = {10.1016/j.eswa.2018.07.070},
	language = {en},
	urldate = {2021-07-23},
	journal = {Expert Systems with Applications},
	author = {Nezhad, Milad Zafar and Sadati, Najibesadat and Yang, Kai and Zhu, Dongxiao},
	month = jan,
	year = {2019},
	pages = {16--26},
}

@article{aastha_deepcompete_2021,
	title = {{DeepCompete} : {A} deep learning approach to competing risks in continuous time domain},
	volume = {2020},
	issn = {1942-597X},
	shorttitle = {{DeepCompete}},
	url = {https://www.ncbi.nlm.nih.gov/pmc/articles/PMC8075516/},
	urldate = {2021-11-17},
	journal = {AMIA Annual Symposium Proceedings},
	author = {Aastha and Huang, Pengyu and Liu, Yan},
	month = jan,
	year = {2021},
	pmid = {33936389},
	pmcid = {PMC8075516},
	pages = {177--186},
}

@inproceedings{zhu_deep_2016,
	address = {Shenzhen, China},
	title = {Deep convolutional neural network for survival analysis with pathological images},
	isbn = {978-1-5090-1611-2},
	url = {http://ieeexplore.ieee.org/document/7822579/},
	doi = {10.1109/BIBM.2016.7822579},
	urldate = {2021-11-30},
	booktitle = {2016 {IEEE} {International} {Conference} on {Bioinformatics} and {Biomedicine} ({BIBM})},
	publisher = {IEEE},
	author = {Zhu, Xinliang and Yao, Jiawen and Huang, Junzhou},
	month = dec,
	year = {2016},
	pages = {544--547},
}

@article{chi_deep_2021,
	title = {Deep {Semisupervised} {Multitask} {Learning} {Model} and {Its} {Interpretability} for {Survival} {Analysis}},
	volume = {25},
	issn = {2168-2194, 2168-2208},
	url = {https://ieeexplore.ieee.org/document/9373895/},
	doi = {10.1109/JBHI.2021.3064696},
	number = {8},
	urldate = {2021-11-30},
	journal = {IEEE Journal of Biomedical and Health Informatics},
	author = {Chi, Shengqiang and Tian, Yu and Wang, Feng and Wang, Yu and Chen, Ming and Li, Jingsong},
	month = aug,
	year = {2021},
	pages = {3185--3196},
}

@article{kim_improved_2020,
	title = {Improved survival analysis by learning shared genomic information from pan-cancer data},
	volume = {36},
	issn = {1367-4803, 1460-2059},
	url = {https://academic.oup.com/bioinformatics/article/36/Supplement_1/i389/5870509},
	doi = {10.1093/bioinformatics/btaa462},
	language = {en},
	number = {Supplement\_1},
	urldate = {2021-12-01},
	journal = {Bioinformatics},
	author = {Kim, Sunkyu and Kim, Keonwoo and Choe, Junseok and Lee, Inggeol and Kang, Jaewoo},
	month = jul,
	year = {2020},
	pages = {i389--i398},
}

@inproceedings{hao_cox-pasnet_2018,
	title = {Cox-{PASNet}: {Pathway}-based {Sparse} {Deep} {Neural} {Network} for {Survival} {Analysis}},
	shorttitle = {Cox-{PASNet}},
	doi = {10.1109/BIBM.2018.8621345},
	booktitle = {2018 {IEEE} {International} {Conference} on {Bioinformatics} and {Biomedicine} ({BIBM})},
	author = {Hao, Jie and Kim, Youngsoon and Mallavarapu, Tejaswini and Oh, Jung Hun and Kang, Mingon},
	month = dec,
	year = {2018},
	pages = {381--386},
}

@article{tang_capsurv_2019,
	title = {{CapSurv}: {Capsule} {Network} for {Survival} {Analysis} {With} {Whole} {Slide} {Pathological} {Images}},
	volume = {7},
	issn = {2169-3536},
	shorttitle = {{CapSurv}},
	doi = {10.1109/ACCESS.2019.2901049},
	journal = {IEEE Access},
	author = {Tang, Bo and Li, Ao and Li, Bin and Wang, Minghui},
	year = {2019},
	note = {Conference Name: IEEE Access},
	pages = {26022--26030},
}

@inproceedings{li_deepcomp_2020,
	title = {{DeepComp}: {Which} {Competing} {Event} {Will} {Hit} the {Patient} {First}?},
	shorttitle = {{DeepComp}},
	doi = {10.1109/BIBM49941.2020.9313333},
	booktitle = {2020 {IEEE} {International} {Conference} on {Bioinformatics} and {Biomedicine} ({BIBM})},
	author = {Li, Yingxue and Jia, Wenxiao and Kang, Yashu and Chen, Tiange and Li, Xiang and Du, Xin and Dong, Jianzeng and Ma, Changsheng and Wang, Fei and Xie, Guotong},
	month = dec,
	year = {2020},
	pages = {629--636},
}

@article{wang_survnet_2021,
	title = {{SurvNet}: {A} {Novel} {Deep} {Neural} {Network} for {Lung} {Cancer} {Survival} {Analysis} {With} {Missing} {Values}},
	volume = {10},
	issn = {2234-943X},
	shorttitle = {{SurvNet}},
	url = {https://www.frontiersin.org/article/10.3389/fonc.2020.588990},
	doi = {10.3389/fonc.2020.588990},
	urldate = {2021-12-02},
	journal = {Frontiers in Oncology},
	author = {Wang, Jianyong and Chen, Nan and Guo, Jixiang and Xu, Xiuyuan and Liu, Lunxu and Yi, Zhang},
	year = {2021},
	pages = {3128},
}

@article{tong_deep_2020,
	title = {Deep learning based feature-level integration of multi-omics data for breast cancer patients survival analysis},
	volume = {20},
	issn = {1472-6947},
	url = {https://bmcmedinformdecismak.biomedcentral.com/articles/10.1186/s12911-020-01225-8},
	doi = {10.1186/s12911-020-01225-8},
	language = {en},
	number = {1},
	urldate = {2021-12-02},
	journal = {BMC Medical Informatics and Decision Making},
	author = {Tong, Li and Mitchel, Jonathan and Chatlin, Kevin and Wang, May D.},
	month = dec,
	year = {2020},
	pages = {225},
}

@article{xie_mixture_2021,
	title = {Mixture cure rate models with neural network estimated nonparametric components},
	volume = {36},
	issn = {0943-4062, 1613-9658},
	url = {https://link.springer.com/10.1007/s00180-021-01086-3},
	doi = {10.1007/s00180-021-01086-3},
	language = {en},
	number = {4},
	urldate = {2021-12-02},
	journal = {Computational Statistics},
	author = {Xie, Yujing and Yu, Zhangsheng},
	month = dec,
	year = {2021},
	pages = {2467--2489},
}

@article{lee_deepbts_2020,
	title = {{DeepBTS}: {Prediction} of {Recurrence}-free {Survival} of {Non}-small {Cell} {Lung} {Cancer} {Using} a {Time}-binned {Deep} {Neural} {Network}},
	volume = {10},
	copyright = {2020 The Author(s)},
	issn = {2045-2322},
	shorttitle = {{DeepBTS}},
	url = {http://www.nature.com/articles/s41598-020-58722-z},
	doi = {10.1038/s41598-020-58722-z},
	language = {en},
	number = {1},
	urldate = {2021-12-06},
	journal = {Scientific Reports},
	author = {Lee, Bora and Chun, Sang Hoon and Hong, Ji Hyung and Woo, In Sook and Kim, Seoree and Jeong, Joon Won and Kim, Jae Jun and Lee, Hyun Woo and Na, Sae Jung and Beck, Kyongmin Sarah and Gil, Bomi and Park, Sungsoo and An, Ho Jung and Ko, Yoon Ho},
	month = feb,
	year = {2020},
	note = {Bandiera\_abtest: a
Cc\_license\_type: cc\_by
Cg\_type: Nature Research Journals
Number: 1
Primary\_atype: Research
Publisher: Nature Publishing Group
Subject\_term: Computational science;Non-small-cell lung cancer
Subject\_term\_id: computational-science;non-small-cell-lung-cancer},
	pages = {1952},
}

@inproceedings{huang_deep_2018,
	address = {Kohala Coast, Hawaii, USA},
	title = {Deep {Integrative} {Analysis} for {Survival} {Prediction}},
	url = {https://www.worldscientific.com/doi/abs/10.1142/9789813235533_0032} ,
	doi = {10.1142/9789813235533_0032},
	language = {en},
	urldate = {2021-12-16},
	booktitle = {Biocomputing 2018},
	publisher = {WORLD SCIENTIFIC},
	author = {Huang, Chenglong and Zhang, Albert and Xiao, Guanghua},
	month = jan,
	year = {2018},
	pages = {343--352},
}

@article{kovalev_survlime_2020,
	title = {{SurvLIME}: {A} method for explaining machine learning survival models},
	volume = {203},
	issn = {0950-7051},
	shorttitle = {{SurvLIME}},
	url = {https://www.sciencedirect.com/science/article/pii/S0950705120304044},
	doi = {10.1016/j.knosys.2020.106164},
	language = {en},
	urldate = {2022-01-12},
	journal = {Knowledge-Based Systems},
	author = {Kovalev, Maxim S. and Utkin, Lev V. and Kasimov, Ernest M.},
	month = sep,
	year = {2020},
	pages = {106164},
}

@inproceedings{haarburger_image-based_2019,
	title = {Image-{Based} {Survival} {Prediction} for {Lung} {Cancer} {Patients} {Using} {CNNS}},
	doi = {10.1109/ISBI.2019.8759499},
	booktitle = {2019 {IEEE} 16th {International} {Symposium} on {Biomedical} {Imaging} ({ISBI} 2019)},
	author = {Haarburger, Christoph and Weitz, Philippe and Rippel, Oliver and Merhof, Dorit},
	month = apr,
	year = {2019},
	note = {ISSN: 1945-8452},
	pages = {1197--1201},
}

@phdthesis{Sonabend2021,
author = {Sonabend, Raphael Edward Benjamin},
pages = {345},
school = {University College London (UCL)},
title = {{A Theoretical and Methodological Framework for Machine Learning in Survival Analysis: Enabling Transparent and Accessible Predictive Modelling on Right-Censored Time-to-Event Data}},
type = {PhD},
url = {https://discovery.ucl.ac.uk/id/eprint/10129352/},
year = {2021}
}

@inproceedings{ranganath2016deep,
  title={Deep survival analysis},
  author={Ranganath, Rajesh and Perotte, Adler and Elhadad, No{\'e}mie and Blei, David},
  booktitle={Machine Learning for Healthcare Conference},
  pages={101--114},
  year={2016},
  organization={PMLR}
}

@article{haider2020effective,
  title={Effective ways to build and evaluate individual survival distributions},
  author={Haider, Humza and Hoehn, Bret and Davis, Sarah and Greiner, Russell},
  journal={The Journal of Machine Learning Research},
  volume={21},
  number={1},
  pages={3289--3351},
  year={2020},
  publisher={JMLRORG}
}

@article{sonabend2022scoring,
  title={Scoring rules in survival analysis},
  author={Sonabend, Raphael},
  journal={arXiv preprint arXiv:2212.05260}       ,
  year={2022}
}

@article{hornik1989multilayer,
  title={Multilayer feedforward networks are universal approximators},
  author={Hornik, Kurt and Stinchcombe, Maxwell and White, Halbert},
  journal={Neural networks},
  volume={2},
  number={5},
  pages={359--366},
  year={1989},
  publisher={Elsevier}
}

@article{harrell1982,
author = {Harrell, Frank E. and Califf, Robert M. and Pryor, David B.},
issn = {0098-7484},
journal = {JAMA},
month = {5},
number = {18},
pages = {2543--2546},
title = {{Evaluating the yield of medical tests}},
url = {http://dx.doi.org/10.1001/jama.1982.03320430047030}           ,
volume = {247},
year = {1982}
}

@article{uno2007,
author = {Uno, Hajime and Cai, Tianxi and Tian, Lu and Wei, L J},
issn = {01621459},
journal = {Journal of the American Statistical Association},
number = {478},
pages = {527--537},
publisher = {[American Statistical Association, Taylor & Francis, Ltd.]},
title = {{Evaluating Prediction Rules for t-Year Survivors with Censored Regression Models}},
url = {http://www.jstor.org/stable/27639883},
volume = {102},
year = {2007}
}

@article{graf1999,
author = {Graf, Erika and Schmoor, Claudia and Sauerbrei, Willi and Schumacher, Martin},
doi = {10.1002/(SICI)1097-0258(19990915/30)18:17/18<2529::AID-SIM274>3.0.CO;2-5},
issn = {0277-6715},
journal = {Statistics in Medicine},
number = {17-18},
pages = {2529--2545},
pmid = {10474158},
title = {{Assessment and comparison of prognostic classification schemes for survival data}},
url = {http://doi.wiley.com/10.1002/%28SICI%291097-0258%2819990915/30%2918%3A17/18%3C2529%3A%3AAID-SIM274%3E3.0.CO%3B2-5}          ,
volume = {18},
year = {1999}
}

@article{fang2021impact,
  title={The impact of training sample size on deep learning-based organ auto-segmentation for head-and-neck patients},
  author={Fang, Yingtao and Wang, Jiazhou and Ou, Xiaomin and Ying, Hongmei and Hu, Chaosu and Zhang, Zhen and Hu, Weigang},
  journal={Physics in Medicine \& Biology},
  volume={66},
  number={18},
  pages={185012},
  year={2021},
  publisher={IOP Publishing}
}

@article{shahinfar2020many,
  title={“How many images do I need?” Understanding how sample size per class affects deep learning model performance metrics for balanced designs in autonomous wildlife monitoring},
  author={Shahinfar, Saleh and Meek, Paul and Falzon, Greg},
  journal={Ecological Informatics},
  volume={57},
  pages={101085},
  year={2020},
  publisher={Elsevier}
}

@article{bice2020deep,
  title={Deep learning-based survival analysis for brain metastasis patients with the national cancer database},
  author={Bice, Noah and Kirby, Neil and Bahr, Tyler and Rasmussen, Karl and Saenz, Daniel and Wagner, Timothy and Papanikolaou, Niko and Fakhreddine, Mohamad},
  journal={Journal of applied clinical medical physics},
  volume={21},
  number={9},
  pages={187--192},
  year={2020},
  publisher={Wiley Online Library}
}

@article{balki2019sample,
  title={Sample-size determination methodologies for machine learning in medical imaging research: a systematic review},
  author={Balki, Indranil and Amirabadi, Afsaneh and Levman, Jacob and Martel, Anne L and Emersic, Ziga and Meden, Blaz and Garcia-Pedrero, Angel and Ramirez, Saul C and Kong, Dehan and Moody, Alan R and others},
  journal={Canadian Association of Radiologists Journal},
  volume={70},
  number={4},
  pages={344--353},
  year={2019},
  publisher={SAGE Publications Sage CA: Los Angeles, CA}
}

@article{vincent2010stacked,
  title={Stacked denoising autoencoders: Learning useful representations in a deep network with a local denoising criterion.},
  author={Vincent, Pascal and Larochelle, Hugo and Lajoie, Isabelle and Bengio, Yoshua and Manzagol, Pierre-Antoine and Bottou, L{\'e}on},
  journal={Journal of machine learning research},
  volume={11},
  number={12},
  year={2010}
}

@inproceedings{Kingma2014,
  author = {Kingma, Diederik P. and Welling, Max},
  booktitle = {2nd International Conference on Learning Representations, {ICLR} 2014, Banff, AB, Canada, April 14-16, 2014, Conference Track Proceedings},
  eprint = {http://arxiv.org/abs/1312.6114v10},
  eprintclass = {stat.ML},
  eprinttype = {arXiv},
  title = {{Auto-Encoding Variational Bayes}},
  year = 2014
}

@article{sonabend.avoiding.2022,
  title = {Avoiding {{C-hacking}} When Evaluating Survival Distribution Predictions with Discrimination Measures},
  author = {Sonabend, Raphael and Bender, Andreas and Vollmer, Sebastian},
  year = {2022},
  month = sep,
  journal = {Bioinformatics},
  volume = {38},
  number = {17},
  pages = {4178--4184},
  issn = {1367-4803},
  doi = {10.1093/bioinformatics/btac451},
  urldate = {2022-11-02},
  note = {https://doi.org/10.1093/bioinformatics/btac451}
}

@article{snell.external.2021,
  title = {External Validation of Clinical Prediction Models: Simulation-Based Sample Size Calculations Were More Reliable than Rules-of-Thumb},
  shorttitle = {External Validation of Clinical Prediction Models},
  author = {Snell, Kym I. E. and Archer, Lucinda and Ensor, Joie and Bonnett, Laura J. and Debray, Thomas P. A. and Phillips, Bob and Collins, Gary S. and Riley, Richard D.},
  year = {2021},
  month = jul,
  journal = {Journal of Clinical Epidemiology},
  volume = {135},
  pages = {79--89},
  issn = {0895-4356},
  doi = {10.1016/j.jclinepi.2021.02.011},
  urldate = {2023-10-04},
  note = {https://www.sciencedirect.com/science/article/pii/S0895435621000482}
}

@article{boulesteix.plea.2013,
  title = {A {{Plea}} for {{Neutral Comparison Studies}} in {{Computational Sciences}}},
  author = {Boulesteix, Anne-Laure and Lauer, Sabine and Eugster, Manuel J. A.},
  year = {2013},
  month = apr,
  journal = {PLOS ONE},
  volume = {8},
  number = {4},
  pages = {e61562},
  publisher = {{Public Library of Science}},
  issn = {1932-6203},
  doi = {10.1371/journal.pone.0061562},
  urldate = {2023-10-04},
  langid = {english},
  note = {https://journals.plos.org/plosone/article?id=10.1371/journal.pone.0061562}
}

@article{niessl.overoptimism.2022,
  title = {Over-Optimism in Benchmark Studies and the Multiplicity of Design and Analysis Options When Interpreting Their Results},
  author = {Nie{\ss}l, Christina and Herrmann, Moritz and Wiedemann, Chiara and Casalicchio, Giuseppe and Boulesteix, Anne-Laure},
  year = {2022},
  journal = {WIREs Data Mining and Knowledge Discovery},
  volume = {12},
  number = {2},
  pages = {e1441},
  issn = {1942-4795},
  doi = {10.1002/widm.1441},
  urldate = {2022-11-22},
  langid = {english},
  note = {https://onlinelibrary.wiley.com/doi/abs/10.1002/widm.1441}  
}

@article{boulesteix.replication.2020,
  title = {A Replication Crisis in Methodological Research?},
  author = {Boulesteix, Anne-Laure and Hoffmann, Sabine and Charlton, Alethea and Seibold, Heidi},
  year = {2020},
  journal = {Significance},
  volume = {17},
  number = {5},
  pages = {18--21},
  issn = {1740-9713},
  doi = {10.1111/1740-9713.01444},
  urldate = {2022-11-22},
  langid = {english},
  note = {https://onlinelibrary.wiley.com/doi/abs/10.1111/1740-9713.01444}  
}

@article{rindt2022,
  title={Survival regression with proper scoring rules and monotonic neural networks},
  author={Rindt, David and Hu, Robert and Steinsaltz, David and Sejdinovic, Dino},
  booktitle={International Conference on Artificial Intelligence and Statistics},
  pages={1190--1205},
  year={2022},
  organization={PMLR}
}

@article{rugamer.new.2023,
  title = {A {{New PHO-rmula}} for {{Improved Performance}} of {{Semi-Structured Networks}}},
  booktitle = {Proceedings of the 40th {{International Conference}} on {{Machine Learning}}},
  author = {R{\"u}gamer, David},
  year = {2023},
  month = jul,
  pages = {29291--29305},
  publisher = {{PMLR}},
  issn = {2640-3498},
  urldate = {2023-10-05},
  langid = {english},
  note = {https://proceedings.mlr.press/v202/rugamer23a.html}
}

\end{document}